\pdfoutput=1
\documentclass[11pt]{article}

\usepackage[preprint]{acl}

\usepackage{times}
\usepackage{latexsym}
\usepackage{array}
\usepackage{multirow}
\usepackage{array}
\usepackage{booktabs}
\usepackage{amsmath} 
\usepackage[T1]{fontenc}
\usepackage{color, colortbl}

\usepackage[utf8]{inputenc}

\usepackage{microtype}

\usepackage{inconsolata}
\usepackage{outlines}
\usepackage{tcolorbox}
\usepackage{mdframed}

\usepackage{amsthm} % for theorem environments
\theoremstyle{definition} % defines the style of theorem environments
\usepackage[textsize=scriptsize]{todonotes}

\title{Molecular Facts: Desiderata for Decontextualization \\in LLM Fact Verification}

\author{Anisha Gunjal \quad\quad\quad Greg Durrett \\
        The University of Texas at Austin \\
        \texttt{anishagunjal@utexas.edu}
}

\begin{document}

\maketitle

\begin{abstract}
Automatic factuality verification of large language model (LLM) generations is becoming more and more widely used to combat hallucinations. A major point of tension in the literature is the granularity of this fact-checking: larger chunks of text are hard to fact-check, but more atomic facts like propositions may lack context to interpret correctly. In this work, we assess the role of context in these atomic facts. We argue that fully atomic facts are not the right representation, and define two criteria for \emph{molecular facts}: decontextuality, or how well they can stand alone, and minimality, or how little extra information is added to achieve decontexuality. We quantify the impact of decontextualization on minimality, then present a baseline methodology \footnote{\url{https://github.com/anisha2102/molecular_facts}} for generating molecular facts automatically, aiming to add the right amount of information. We compare against various methods of decontextualization and find that molecular facts balance minimality with fact verification accuracy in ambiguous settings. 

%\gd{I rewrote this last sent}
%We examine the impact of decontextualization on minimality through a controlled experiment, identifying when decontextualization adds too much information and makes generated atomic facts ineffective for error localization. 
% \ag{update this} molecular facts reduce false verification rates drastically for ambiguous claims, particularly when multiple evidence documents contain confusing entities sharing similar names.
%\gd{I removed 6\%, needs context from later}

% - analyse decontextualization (how much is good/how much is bad)\\
% - promising future directions for RAG systems/Commercial systems like perplexity/bing

\end{abstract}

\section{Introduction}

 %for generating content through parametric memory and retrieval-aided generation. While these models are lauded for their ability to generate coherent and informative content, they are also prone to generating unfaithful or hallucinated content \cite{ji2022survey, zhang2023siren}. Moreover, responses from generative search engines like BingChat and perplexity.ai further exemplify this issue, frequently delivering outputs that, despite their fluent and seemingly informative nature, often contain factually unsupported claims and inaccurate references \cite{liu2023evaluating}. 
Large language models (LLMs) have emerged as powerful tools for delivering knowledge to users, either via closed-book generation or retrieval-augmented systems. However, these systems may not always produce correct facts \cite{liu2023evaluating}, an instance of the ``hallucination'' problem \cite{zhang2024how, ji2022survey, zhang2023siren}. Recent research has shown the potential of LLMs to identify unfaithful content and enable automatic fact-checking and attribution against sources \cite{falke2019ranking,goyal2020annotating,min-etal-2023-factscore, wang2024factcheckbench, chern2023factool, wei2024long, chen2023felm, malaviya2023expertqa, gao2023enabling, tang2024minicheck}.
%\gd{I would flesh this out in related work, maybe add Tanya NAACL 2021 papers and Falke 2019} \ag{added liyan, not sure which one from tanya, and yet to add pre 2023}\gd{updated} %These studies propose methodologies aimed at streamlining the fact-checking process for LLM-generated content by integrating retrieval mechanisms and minimizing human intervention.

%This typically includes breaking down the generated content into individual atomic claims, assessing these for checkworthiness, revising them, and validating their accuracy against retrieved evidence.
\begin{figure}
    \centering
    \includegraphics[scale=0.73,trim=30mm 115mm 100mm 5mm]{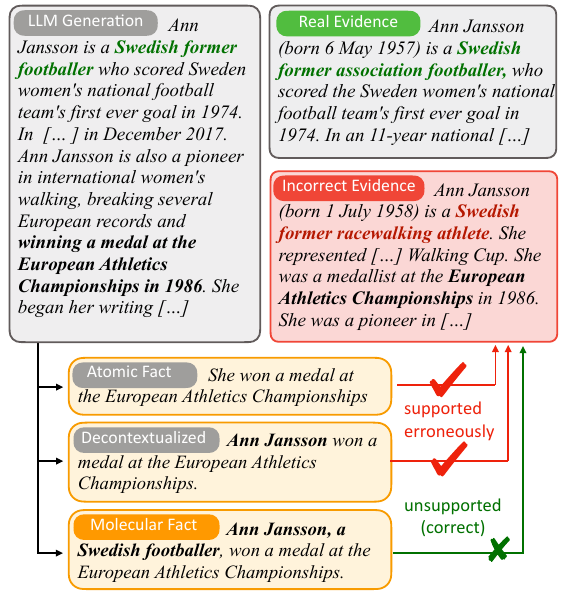}
    \caption{Breaking a paragraph into atomic facts can cause errors in attribution: facts out of context appear to be true when they are not. The right granularity of decontextualization, ``molecular facts,'' balances contextual grounding with atomicity.}
    \label{fig:intro}
\end{figure}

% \begin{figure}
%     \centering
%     \includegraphics[scale=0.75,trim=40mm 105mm 100mm 15mm]{figs/intro.pdf}
%     \caption{Breaking a paragraph into ``atomic facts'' can cause errors in attribution: facts out of context appear to be true when they are not. The right granularity of decontextualization, ``molecular facts,'' balances contextual grounding with atomicity.}
%     \label{fig:intro}
% \end{figure}
% resolved \greg{example is not as clear as it could be, that evidence is retrieved and not about the same person as the generation...maybe need to show two evidence paragraphs. maybe even have the atomic fact be ``she'' and then show ``Ann Jansson'' and ``Ann Jansson, a Swedish footballer'' as the two augmentations?}

% \gd{cite qafacteval, sihao chen propsegment}
% \gd{cite rarr, jifan naacl '24, factcheck-gpt}
A key step in this process is to break down generated content into individual atomic claims \cite{fabbri-etal-2022-qafacteval, chen-etal-2023-propsegment,kamoi-etal-2023-wice,min-etal-2023-factscore}. This decomposition allows for retrieval of evidence focused on a particular part of the generated content \cite{gao-etal-2023-rarr,wang2024factcheckbench, chen2023complex} and also error localization by determining which parts of the content are supported or not. However, this step is not straightforward. \citet{wanner2024closer} highlights that the effectiveness of automatic factuality verification is heavily dependent on the strategies employed for decomposing content into claims. In particular, LLMs have a propensity to incorrectly merge information about similarly named entities \citet{lee2024ambigdocs} and current evaluation methods struggle to handle these ambiguities in atomic claims \cite{chiang2024merging}. Figure~\ref{fig:intro} shows a possible issue: a fact that is ``too atomic'' can be validated against evidence that doesn't actually support it.

%Furthermore, \citet{lee2024ambigdocs} show that LLMs have a propensity to incorrectly merge information about similarly named entities.
%While atomic claims do simplify the evaluation process and improve error localization for complex claim verification , it also introduces the risk of diminished precision due to contextual misunderstandings and ambiguities. This raises a crucial question: How do we determine the the optimal framing of claims for factuality verification?

%\greg{comment on the example}
In this work, we address the problem of how to find minimal yet still unambiguous facts for LLM fact verification. We frame this problem as one of \emph{decontextualization}, adding context to a sentence to make it stand alone while retaining its original meaning \cite{choi2021decontextualization}. This process draws on the idea of \textit{specificity} from discourse \cite{louis2012corpus}, specifically whether sentences can express key information about the participants without ambiguity \cite{li2016improving}. %Recent work in decontextualization has gone beyond simple anaphora resolution with techniques like rewriting sentences using a question-answering framework \cite{newman2023question}. We go one step further:
However, making a claim unambigous is not enough: when escalating from simple pronoun replacement in atomic facts to elaborations like \emph{a Swedish footballer} in Figure~\ref{fig:intro}, we must balance the specificity of the fact with how easy it will be to verify. It is not trivial to select the ``right'' information to elaborate on a claim without compromising the ease of verification. 

We define two criteria needed in this fact-checking setting: \textit{decontextuality}, where the claim should uniquely specifying entities, events, and context, and \textit{minimality}, maintained by avoiding excessive additional information that could complicate verification. We propose a notion of \emph{molecular facts}, which balances these two criteria: molecular facts should be fully specific while compatible with the maximum number of possible evidence documents. We explore these criteria and our molecular facts in two settings. First, we address the question of how much \emph{non-minimality} could be a problem for error localization with standard decontextualization techniques. We devise a synthetic fact-checking experiment where particular nuances of an output generation are unsupported and show that an average of 6\% of claims may pose problems for error localization. In a setting with LLM responses of 5 sentences with 3 claims each, this would lead to localization errors in a large fraction of responses. We then evaluate the opposite problem, whether decontextualization is \emph{too minimal}. We study a dataset of fact-checking with ambiguous entity names presented in \citet{chiang2024merging}. We show that our method of molecular fact generation balances accuracy under ambiguous entity references with minimality of claims.
% past approaches to decontextualization for claims with ambiguous entity references while maintaining minimality. %Ultimately, our results indicate that balancing the granularity of decontextualization is important for a system to navigate both settings successfully. 
%\gd{one thing I think we don't say in the intro is that it's not just the amount of excess information, but how easy/hard it is to verify...we may want to say something about this} 
%\greg{summarize results here}
%\gd{Anisha double check this claim}
% \gd{revisit and make this statement crisper once results are more clear} 

% \anisha{refine this after final pass}
Our main contributions are: (1) We re-examine the decontextualization process for fact-checking and define \emph{molecular claims} following the desiderata of decontextuality and minimality. (2) We investigate the loss of minimality due to claim decontextualization and its impacts on error localization. (3) We find that molecular claims are more performant and minimal for long-form generations than existing decontextualization methods.

\section{Desiderata for Decontextualization}

%\gd{I think we need to acknowledge that atomic is not really precise. you can very frequently make more facts or split even smaller} \ag{modified}
We propose desiderata to determine the optimal level of decontextualization required for atomic facts. An \emph{atomic fact} is defined as a discrete unit of information, derived from a broader claim, and variously described in the literature as propositions, subclaims, summary content units, or atomic content units \cite{nenkova2004evaluating, liu-etal-2023-revisiting, zhang-bansal-2021-finding, chen-etal-2023-propsegment, min-etal-2023-factscore, kamoi-etal-2023-wice}. 
%resolved \gd{add citations} 

%done \greg{cite Ido Dagan and Dan Roth stuff} 
% \gd{maybe pick different example}
Although an atomic fact theoretically represents a singular conceptual unit, recent NLP work using this does not typically give this a rigorous definition from the standpoint of semantics. %Facts are often equated with propositions \cite{ernst2021proposition, chen-etal-2023-propsegment}, but propositions are not necessarily the right granularity: for example, \emph{the president gave the speech to Congress on Thursday} has several units to check (was the speech directed at Congress? did it take place on Thursday?). As a result, different work takes different granularity of atomic facts regardless. 
\citet{wanner2024closer} demonstrate a high variation in the number of subclaims generated by different decomposition methods, with the macro-average of subclaims per biography ranging from 20.2 using the method by \citet{kamoi-etal-2023-wice} to 32.9 with the approach by \citet{chen-etal-2023-propsegment}. Note that in Figure~\ref{fig:intro}, \emph{She was a medallist at the European Athletics Championships in 1986} could be kept as one unit or broken into three facts evaluating her status as a medallist, the venue, and the date.

\subsection{Desiderata}
\label{sec:desiderata}
\paragraph{Preliminaries} We define $\mathbf{r}$ as a response from a language model to an input prompt $\mathbf{x}$, consisting of a series of claims $(\mathbf{c}_1,\ldots,\mathbf{c}_n)$ to be verified. Claims are extracted through an upstream process of decomposition and potentially filtering for ``check-worthiness'' (i.e., does the claim present factual content or does it present an opinion?). We describe the prompting in Appendix \ref{appendix:prompts}. 

% \ag{commented ${r}_{<i}$ }
%Define $\mathbf{r}_{<i}$ to be the response \emph{context} occurring prior to claim $\mathbf{c}_i$. For example, if a claim is extracted from the fourth sentence of a response, then $\mathbf{r}_{<i}$ should contain the first three sentences and any necessary parts of the fourth, since these could include referents of expressions in $\mathbf{c}_i$.

%\gd{reference appendix} 

We assume that in the context of $\mathbf{r}$ and $\mathbf{x}$, a claim $\mathbf{c}_i$ can be fully interpreted with a truth-conditional meaning $I(\mathbf{c}_i \mid \mathbf{x}, \mathbf{r})$. In the terminology of \citet{Rashkin2021MeasuringAI} and \citet{choi2021decontextualization}, $I(\mathbf{c}_i \mid \mathbf{x}, \mathbf{r})$ represents $\mathbf{c}_i$ interpreted in the \emph{linguistic context} of $\mathbf{x}$ and $\mathbf{r}$.

% We assume that in the context of $\mathbf{r}$ and $\mathbf{x}$, a claim $\mathbf{c}_i$ can be fully interpreted with a truth-conditional meaning $I(\mathbf{c}_i \mid \mathbf{x}, \mathbf{r}_{<i})$. In the terminology of \citet{Rashkin2021MeasuringAI} and \citet{choi2021decontextualization}, $I(\mathbf{c}_i \mid \mathbf{x}, \mathbf{r}_{<i})$ represents $\mathbf{c}_i$ interpreted in the \emph{linguistic context} of $\mathbf{x}$ and $\mathbf{r}$. 

%We treat $I$ as purely abstract in this work; we will not further flesh out its semantics. %Instead, we can think of $I$ at a high level in two ways. %First, $I$ could simply be the string $(\mathbf{x}, \mathbf{r}, \mathbf{c})$ concatenated together, with the meaning ``interpret $\mathbf{c}$ in this discourse context''.
%\jl{reaction at this point, maybe I'm not understanding things: why is $r$ necessary for the interpretation? And if $c$ is just a set of claims and $x$ consists of $c$, how could a complete discourse interpretation come about unless every sentence in a document is a claim (but that's not the case since things are already filtered)?}
We can construct a \emph{standalone proposition} with truth conditional meaning equivalent to $I$ by being sufficiently specific. For example, the statement in Figure~\ref{fig:intro} could be completely specified as \emph{Ann Jansson, the Swedish footballer born on 6 May 1957 who played for Hammarby IF, won a medal at the European Athletics Championship, the biennial event organized by the European Athletics Association, in 1986.} %In this sentence, there is no doubt about which athlete and what event are being referred to.%\jl{Not sure what you're trying to get at... What's the difference between these two interpretations?} \ag{TODO: make this less confusing}

\paragraph{Decontextualization} Our goal in this work is to produce rewritten \emph{molecular claims}. Denote by $\mathbf{m}_i$ the rewritten form of $\mathbf{c}_i$, which should have semantics $I$ when interpreted as a standalone proposition.
%Our goal in this work is to produce a representation of a claim \ag{M is the output representation of the claim by this method}  $M(\mathbf{c}, \mathbf{x}, \mathbf{r}) = \mathbf{c} + \mathrm{Dis}(\mathbf{c}, \mathbf{x}, \mathbf{r})$\jl{Wait why is M the claim now, aren't $c$'s claims? Also what is $r$ doing here?} that is equivalent to $I$. This definition should be read in terms of the information content of $M$: it contains the information in $\mathbf{c}$ as well as disambiguating information $\mathrm{Dis}$ which elaborates on the claim.
As in Figure~\ref{fig:intro}, this requires adding disambiguating information that could provide information needed to identify an entity (specifying that Jansson is a \emph{Swedish footballer}), identify an event (specifying that the event happened in 1986), specify a qualification (\emph{in the field of biochemistry, ...}), or more.

%Let $E(\mathbf{c})$ denote the entities associated with a claim.

%Let \( E \), denoted as \( C(E) \); the disambiguating information, \( \text{Dis}(E) \); and the contextual information about the entity from the long-form content, \( \text{Context}(E) \). The disambiguating information \( \text{Dis}(E) \) includes specific details that clarify and distinguish \( E \) from other similarly named entities, such as profession, location, or historical context. The contextual information, \( \text{Context}(E) \), adds further details relevant to the entity derived from the generated content.\gd{I read up to here}

%The final integrated statement \( M(E) \), which incorporates the claim, disambiguating information, and context, is defined as:
%\[
%M(E) = C(E) + \textit{Dis}(E) + \textit{Context}(E)
%\]
%This equation represents the incorporation of necessary pertinent information into the core atomic fact to effectively reduce ambiguity regarding the entity \( E \). We further conceptualize the following criteria for the selection of the disambiguation information Dis(E).

% \begin{tcolorbox}[title={Criterion 1}]
%     Disambiguation of the entity should provide a \textbf{uniquely identifying} property of an entity against other entities with the same name.
% \end{tcolorbox}

\paragraph{\colorbox[HTML]{DCE5D2}{Criterion 1 (Decontextuality)}}
    \textit{When interpreted as a standalone statement, $\mathbf{m}_i$ must have the truth conditional meaning $I(\mathbf{c}_i, \mathbf{x}, \mathbf{r})$. That is, it should uniquely specify entities, events, and other context such that the claim $\mathbf{c}_i$ is now interpretable.}

%\ag{resolved} \gd{added at least, check wiki, there are many}

This criterion is equivalent to Definition 1 from \citet{choi2021decontextualization}. For the settings we consider, the level of added information needed to specify the meaning of a statement like that in Figure~\ref{fig:intro} may be higher than in past applications like \citet{choi2021decontextualization}. % can face challenges when the type of disambiguation required for a claim is only evident when considering broader context or general world knowledge. \jl{So the Newman paper https://arxiv.org/abs/2305.14772 definitely does not suffer from this issue.} \ag{My take on this, based on my first skim of the paper today, is that it's difficult for Newman style method to get to these cases as they generate question by only seeing the claim, so if the questions don't target things that are present in the context, it may get missed by their rewriting step. I tried this exact example with their prompt, and it doesn't comes up with a question like what is the profession of Ann Jansson, but more like (What event did Ann Jansson compete in / which year/ which medal) }
It is not sufficient to replace the pronoun \emph{she} with \emph{Ann Jansson}; we need to specify \emph{Ann Jansson, the Swedish footballer}. Similarly, the city \emph{George Town} could refer to a city in the Cayman Islands or Malaysia, therefore it must be decontextualized appropriately with a descriptor like \emph{George Town, a city in Cayman Islands}.

Other work such as question answering frameworks based on clarifying questions can target this information \cite{newman2023question}, but may fail to integrate the minimal new information needed, which we describe next.

%This criteria implies that \( \mathrm{Dis} \) must be chosen carefully

%should be chosen such that it minimizes fraction of entities named E that are consistent with \( \text{Dis}(E) \). For example, the entity `George Town' can be described in various ways such as `the capital city', `a city in the Cayman Islands', `a city in the heart of the Caribbean', or `the largest urban center'. Given the ambiguity—since `George Town' refers to locations in at least two locations, the Cayman Islands and Malaysia—it becomes crucial to incorporate a descriptor that clearly identifies the intended George Town. Therefore, to disambiguate, it is necessary to include a contextual cue indicating the affiliating country or colony. For instance, the contextual cue `George Town is a city in Cayman Islands' provides clarity by distinguishing it from other entities with the same name.

% \gd{I don't get how criterion 2 isn't implied by 1 and 3? if it uniquely identifies the entity + preserves the association, that's fine, right?}
% \ag{Removed criterion 3 as discussed in last meeting}

\paragraph{Minimality} Adding too much information to a claim makes it less minimal. For instance, replacing ``\emph{Ann Jansson}'' with ``\emph{Ann Jansson, a Swedish footballer}'' requires verifying that a context referring to Ann Jansson is indeed talking about the Swedish footballer. Taken further, the reference ``\emph{Ann Jansson, the Swedish footballer born on 6 May 1957 who played for Hammarby IF}'' is clearly suboptimal. It requires verifying Jansson's birthdate as an additional detail, and crucially, this detail won't be frequently reported in documents about Ann Jansson.

Define $\mathcal{E}^*(I(\mathbf{c}, \mathbf{x}, \mathbf{r}))$ as the set of set of evidence documents that support the statement $I$ with an \emph{oracle} understanding of the entities involved. For instance, this would contain a document describing the correct Ann Jansson, even if it did not confirm all the details about her life. Define $\mathcal{E}(\mathbf{m}_i) \subset \mathcal{E}^*$ to be the set of evidence documents that fully support a statement $\mathbf{m}_i$. For instance, in the case of Ann Jansson above, the document would need to specify Jansson's birthdate if this is contained in $\mathbf{m}$. %; we would not assume a separate step where we can validate this information from Wikipedia.

%Our second criterion follows from the desire to preserve as many candidate documents as possible for the set $\mathcal{E}$.

\paragraph{\colorbox[HTML]{DCE5D2}{Criterion 2 (Minimality)}}
    \textit{Given a set of statements $\mathcal{M}$ that all decontextualize a claim $\mathbf{c}_i$, we should select $\mathrm{argmax}_{\mathbf{m} \in \mathcal{M}} | \mathcal{E}(\mathbf{m})| $ to maximize the size of the set of supporting evidence documents.}

This criterion means that, when selecting distinguishing details for an entity, we should choose those that can typically be inferred from evidence. For instance, ``\emph{Jason Martin}'' may be characterized either as a ``\emph{rugby player}'' or specifically as a ``\emph{former player for North Queensland Cowboys}.'' Since ``\emph{rugby player}'' is a more enduring and widely recognized description, yet still specific enough to indicate Jason Martin, it is more likely to be supported by a larger number of documents.

Past work like \citet{choi2021decontextualization} instructs annotators to make minimal edits to statements. However, they do not provide guidance on what criteria should be used to choose from among multiple candidate edits.
\paragraph{Molecular facts} These two criteria suggest two things. First, atomic facts can be ``too atomic:'' they may need to be decontextualized. However, it is still valuable to have a reasonably minimal fact so it can be supported by many possible evidence documents.

%Bridging these two criteria, we set a clear standard for refining information about an entity \( E \). The first criterion focuses on uniqueness; any additional detail \( \text{Dis}(E) \) we provide must clearly single out \( E \) from others entities with the same name. This is about ensuring precision of evaluating claims about the right unambiguous entity. The second criterion emphasizes commonality—the details we add should be those most likely found in documents about \( E \). This is about preserving the recall and ensuring that fact verification is not impacted due to introduction of uncommon facts that will not be present in majority documents of a factual claim's evidence base. Together, these criteria guide us in crafting a 'molecular fact': a statement that accurately identifies \( E \) and is backed by the majority of evidence, striking a balance between being exact and being commonly recognized.

\begin{figure*}
    \centering
    \includegraphics[scale=1,trim=40mm 125mm 90mm 35mm]{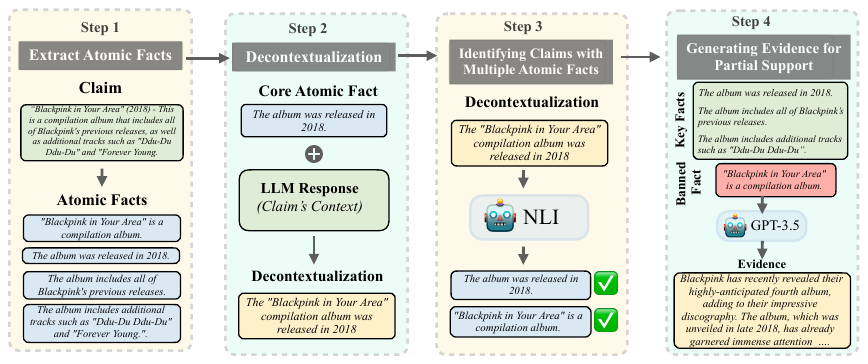}
    \caption{Controlled evidence generation framework for illustrating error localization introduced by decontextualization for atomic fact verification.}
    \label{fig:error_localization}
\end{figure*}

\paragraph{\colorbox[HTML]{F7EBCC}{Molecular Fact}} \textit{A molecular fact is a statement $\mathbf{m}_i$ corresponding to claim $\mathbf{c}_i$ that obeys criteria 1 and 2: it should uniquely specify the interpretation of $\mathbf{c}_i$ even when considered on its own, while adding as little information as possible to do so.}

%To ensure the preservation of atomicity while localizing errors within complex claims, a molecular fact should be as close as possible to the atomic fact. However, it must also include the necessary critical information to disambiguate the entities within the atomic claim.

% \greg{Suppose we have entity E with claim C(E). We want to produce a new statement that conveys both C(E) and Dis(E), where Dis(E) is information that disambiguates E. Dis(E) should be inferable from the context Context(C), and should maximize: fraction of evidence docs associated with E that also specify Dis(E), how uniquely it specifies E (minimize fraction of things named E consistent with Dis(E))} \ag{Incorporated this above}

\subsection{Task Definition: Fact-checking LLMs}
\label{sec:task}
%$\mathbf{r} = (\mathbf{c}_1,\ldots,\mathbf{c}_n)$ 
% \gd{refactor to use the x/r/c notation from sec 2}
%\gd{this needs to be unified with the notation at the start of 2.2}
%We consider a long-form response generated by a large language model, represented by $\mathbf{r}$, consisting of a series of atomic facts, denoted as $(\mathbf{c}_1,\ldots,\mathbf{c}_n)$. These atomic facts are the fundamental elements of information extracted from the response, although they may lack comprehensive context or specific disambiguating details.
Recall our setting where an LLM has generated a response $\mathbf{r}$ to input prompt $\mathbf{x}$, and $\mathbf{r}$ has associated claims $(\mathbf{c}_1,\ldots,\mathbf{c}_n)$.
For each $\mathbf{c}_i$, we have a corresponding set of $k$ evidence documents, \( D_{i} =(D_{i,1},\ldots ,D_{i,k}) \), that are referenced to assess the accuracy of $\mathbf{c}_i$. Furthermore, we have access to a gold standard of human-annotated labels for each atomic fact, represented as \( L = (l_1,\ldots,l_n) \), where each \( l_i \) can be either \texttt{\small{SUPPORTED}} or \texttt{\small{NOT\_SUPPORTED}}. \textbf{Our goal is to make judgments about the supportedness of the  $\mathbf{c}_i$}, which requires appropriately decontextualizing each fact.

We augment each atomic claim $\mathbf{c}_i$ to a corresponding molecular claim $\mathbf{m}_i$ as described in Section~\ref{sec:methodology}, resulting in a set of facts $(\mathbf{m}_1,\ldots{\mathbf{m}_n})$. We represent the model's factuality judgment prediction as a set of supported documents ${p}_i =  \mathrm{Check}(D_{i,j}, \mathbf{m}_i), \quad \text{for all } j \in \{1, 2, \dots, n\} \text{ in }D_{i}$. In other words, the prediction of $\mathrm{Check()}$ is accurate when it supports the molecular claim with the same evidence docs as humans.

%These labels are determined by considering each atomic fact within the broader context of the full LLM response. This means that humans consider the entire generation from the LLM as a contextual backdrop for each claim, incorporating any relevant auxiliary information it provides to make their judgments regarding factuality.

\section{Method: Producing Molecular Facts}
\label{sec:methodology}
We use a two-step process to refine an atomic fact into a molecular fact using \texttt{\small gpt-4-turbo-2024-04-09} \cite{achiam2023gpt}. Our methodology makes the assumption that the ambiguity is typically restricted to a single entity in the claim. This is the case for the datasets we study in this work, described in Section~\ref{sec:setup}. 

%\gd{I just removed the confusing comments here} % (e.g., \emph{Ann Jansson was born in Sweden} relies on disambiguating \emph{Ann Jansson}, but \emph{Sweden} is not ambiguous). rarely has additional ambiguity. \jl{I don't get what you mean by ``additional ambiguity''} \ag{The fact chosen for disambiguation rarely has more ambiguities because it contains the exact information that's necessary to disambiguate the entity. For example, if an entity "X" is ambiguous, if the fact chosen for disambiguation is "X is born in US" then this chosen fact is usually unambiguous. }

\paragraph{\colorbox[HTML]{D3DEE1}{Stage 1: Identifying Ambiguity}}
We identify the primary subject of the claim and to assess potential ambiguities based on its parametric knowledge: does the model know of multiple entities with this name?
%\gd{removed the example, was kind of long-winded and it's covered in the next paragraph} %Some strings, like ``\emph{Charles Osgood}'', refer to multiple entities; e.g., Charles Osgood the writer and commentator (1933–2024), the artist (1809–1890), and the psychologist (1916–1991). On the other hand, certain names, like ``\emph{Julius Robert Oppenheimer}'', are inherently distinct, thus negating the need for additional disambiguation.
This step identifies the main subject $\mathbf{s}_i$ of the claim $\mathbf{c}_i$ and provides a disambiguation criteria $\mathbf{b}_i$ for the subject $\mathbf{s}_i$. The disambiguation criteria $\mathbf{b}_i$ can be `None` when there is no ambiguity, or a type of criteria such as profession, birthyear, or location when disambiguation is required. 

For example, if the claim is about `Charles Osgood', with multiple possible referents, $\mathbf{s}_i$ is \textit{`Charles Osgood'}, while $\mathbf{b}_i$ could be \textit{`profession'} or \textit{`birthyear'} to clarify which Charles Osgood is being referred to. Conversely, if the claim concerns the unambiguous `\textit{Julius Robert Oppenheimer'}, $\mathbf{s}_i$ is \textit{`Julius Robert Oppenheimer'}, and $\mathbf{b}_i$ is \textit{`None'}.
% \gd{please define notation about what the outcome of this stage is. do you identify in a binary fashion when additional context is needed?}

\paragraph{\colorbox[HTML]{D3DEE1}{Stage 2: Molecular Facts Generation}} We then prompt the LLM to disambiguate the subject $\mathbf{s}_i$  within the claim $\mathbf{c}_i$, harnessing both the identified disambiguation criteria $\mathbf{b}_i$ and the claim's context $\mathbf{r}$.%\gd{changed this} \ag{as this data doesn't provide sentence level claims splits, we give the entire response as the context, should I modify this even in the definition above? }\gd{yeah then just give all of r and not $r_{<i}$} %The set $(\mathbf{s_i}, \mathbf{b}_i,\mathbf{r}_{<i})$ is our proposed methodology to infer $\mathrm{Dis}(\mathbf{c}, \mathbf{x}, \mathbf{r})$ for generating molecular fact for a claim $\mathbf{c_i}$ as described in  Section~\ref{sec:desiderata}.
The output of this stage is a molecular fact $\mathbf{m}_i$ for the atomic claim $\mathbf{c}_i$.%\gd{I commented out everything else because I didn't understand it}

The specifics regarding the prompts used are elaborated upon in Appendix \ref{fig:ambig_prompt} and \ref{fig:molec_prompt}.

\subsection{Baselines}
\label{sec:baselines}

% \ag{moved baselines here because we have two baselines in the section of minimality in comparison to the thesis verison}
We analyze the robustness of fact verification across various systems on the defined criteria of \textit{minimality} and \textit{decontextuality}. Outputs for baselines are generated with \texttt{\small{gpt-4-turbo-2024-04-09}}.

\begin{description}
    \item[\texttt{\small{ATOMIC}}:] Atomic claims are generated from the LLM's response using \citet{min-etal-2023-factscore}. \vspace{-0.08in}
    % We reuse the pre-existing atomic claims extracted using the method in \citet{min-etal-2023-factscore} from the ambiguous biographies dataset \cite{chiang2024merging}.
    \item[\texttt{\small{SIMPLE-DECONTEXT}}:] Atomic claims are decontextualized with a prompt described in \ref{fig:decontext_prompt} using the LLM's generated response as context for the atomic claim.\vspace{-0.08in}
    \item[\texttt{\small{SAFE-DECONTEXT}}:] Decontextualization of atomic claims is performed using the revision prompt described in \citet{wei2024long}.\vspace{-0.08in}
    \item[\texttt{\small{MOLECULAR-DECONTEXT}}:] This approach follows a two-stage process described in section \ref{sec:methodology} to identify disambiguation criteria and subsequently decontextualize the atomic claim. \vspace{-0.08in}
\end{description}

\begin{figure*}
    \centering
    \includegraphics[scale=0.75,trim=40mm 170mm 40mm 20mm]{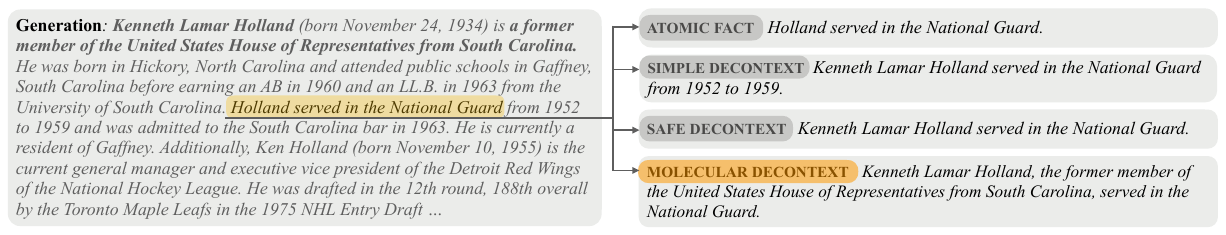}
    \caption{Example claims (right) generated by \texttt{\small{SIMPLE-DECONTEXT, SAFE-DECONTEXT, MOLECULAR-DECONTEXT}} for the atomic claim derived from the highlighted sentence in the LLM generation (left). }
    \label{fig:example-dataset}
\end{figure*}

Examples of outputs from each method can be found in Figure \ref{fig:example-dataset}. With this task definition and baseline methodologies, we structure our experiments to analyze the two criteria presented in Section \ref{sec:desiderata} in the following sections.

% \subsection{ \colorbox[HTML]{F7EBCC}{Minimality Investigation}}
\section{Experiment: Minimality \& Localization} 
We begin our analysis of decontextualization with a controlled experiment to illustrate problems with error localization due to loss of minimality discussed in Criterion 2 in Section~\ref{sec:desiderata}. Minimality is more difficult to evaluate than decontextuality. Less minimal facts impact error localization and can potentially lead to errors where an ancillary part of the claim leads to the whole claim being judged as wrong \cite{kamoi2023shortcomings}. However, precisely measuring the harms of this is not easy without taking into account the downstream uses of error localization systems such as answer refinement \cite{xu2023pinpoint} or fine-tuning \cite{wu2024fine, roit2023factually}. 
%\gd{cite fine-grained RLHF, and maybe the Paul Roit fine-tuning for entailment stuff}
%\gd{cite LLMRefine}  

To measure the effects in a controlled way, we design a method for synthetic evidence generation as summarized in Figure \ref{fig:error_localization}. \textbf{Our goal is to illustrate when decontextualized atomic facts actually contain \emph{multiple} facts in a way that could impact error localization.} We then study how many of these cases truly show this problem. To study the impact of information addition, we consider two baselines \texttt{\small{SIMPLE-DECONTEXT}} and \texttt{\small{SAFE-DECONTEXT}} which respectively have less and more restrictive prompts for including new information from the context to revise an atomic claim.  %Our goal is to analyze the impacts of merged facts due to decontextualization on the localization of errors in LLM generations. We do this by artificially controlling which facts are and aren't supported by a context and using that to assess different decontextualization strategies.
%\jl{The takeaways seem to be that only 6\% of the data has this problem --- so it isn't much of a problem? But is this what you are trying to establish?}

% \notes{Previous works advocate atomicity without merging multiple facts, but in some cases it is essential.}

\subsection{Controlled Dataset Construction} 
We now detail the dataset construction process as illustrated in Figure~\ref{fig:error_localization}. We take a dataset $D$ of 812 claims from the Factcheck-Bench dataset \cite{wang2024factcheckbench} which consists of long form ChatGPT responses with human-annotated factuality labels.

\textbf{Step 1: Extract Atomic Facts} For each response $\mathbf{r} \in D$, we extract atomic facts $(\mathbf{c}_1,\ldots,\mathbf{c}_n)$ using the method of \citet{min-etal-2023-factscore}. 

\textbf{Step 2: Decontextualization:} We perform decontextualization of the extracted atomic facts using \texttt{\small{SIMPLE-DECONTEXT}} and \texttt{\small{SAFE-DECONTEXT}}. Let the decontextualization for claim $\mathbf{c}_i$ be denoted as $\mathbf{d}_i$. We refer to the $\mathbf{c}_i$ that $\mathbf{d}_i$ was created from as its \emph{core atomic fact}; however, note that $\mathbf{d}_i$ might support other facts as well.
%\gd{make clear this is with GPT-3.5} 

\textbf{Step 3: Identifying Claims with Multiple Atomic Facts:} We identify decontextualized claims that entail information of more than one atomic fact. We use the entailment model from \citet{liu2022wanli} to determine $e(\mathbf{d}_i, \mathbf{c}_j) \in \{\mathrm{supported},\mathrm{unsupported}\}$; is each $\mathbf{c}_j$ supported by $\mathbf{d}_i$? %decontextualization to its entailed atomic facts resulting in 
%$\mathbf{d}_i \to [\mathbf{c}_a, \mathbf{c}_b, \ldots, \mathbf{c}_n]$ where \( n \geq 0 \). For our generation, we require that a decontextualization $\mathbf{d}_i$'s corresponding claim $c_i$ is the core key fact and must be present in its mapping set. First, we filter the atomic facts mapped to a decontextualization that are similar to each other (if atomic fact is a substring of another, we retain the larger one). Next, we filter this subset to select the decontextualized facts that are mapped to at least 2 atomic facts i.e. \( n \geq 3 \). These steps lead to a filtering out of 54.77\% of the claims. Let's call this filtered set as $D'$. 
We retain cases where $e(\mathbf{d}_i, \mathbf{c}_i) = \mathrm{supported}$ %; that is, $\mathbf{d}$ supports its corresponding $\mathbf{c}$,
and where $|\{j: e(\mathbf{d}_i, \mathbf{c}_j) = \mathrm{supported}\}| \geq 2$; that is, at least two atomic facts are supported by $\mathbf{d}_i$. For example, in Figure \ref{fig:error_localization}, the claim ($\mathbf{d}_i$), \textit{`The ``Blackpink in Your Area'' compilation album was released in 2018`}, is a decontextualized claim derived from the core atomic claim ($\mathbf{c}_i$), \textit{`The album was released in 2018.'}. The decontextualized claim ($\mathbf{d}_i$) entails the core atomic fact ($\mathbf{c}_i$) and an additional atomic fact ($\mathbf{c}_j$) \textit{` ``Blackpink in Your Area'' is a compilation album'}. Let $D'$ denote this filtered set.
%\greg{Reference a figure here. 
% These steps lead to a filtering out of 54.77\% of the claims. Let's call this filtered set as $D'$. 

% This step leads to filtering out 33.95\% of the claims. 

\label{subsec:step4}
\textbf{Step 4: Generating Evidence for Partial Support:} Whenever multiple atomic facts are merged, we could \emph{theoretically} see a loss in localization capability from a model: if one fact is not supported, the entire claim will be determined to be not supported. To demonstrate this possibility, we now \textbf{generate} evidence that partially supports our multi-fact claims. %, as a proof-of-concept that such evidence could be found in the wild.
%\gd{moved this earlier} 
As an example, in Figure \ref{fig:error_localization}, our goal in step 4 is to generate a paragraph that \emph{should not} include details about ``Blackpink in Your Area'' being a compilation album. Then, if the statement \textit{`The album was released in 2018'} is decontextualized to include information about it being a compilation album, this paragraph will enable us to identify this: the evidence will no longer support the decontextualized fact, reflecting a failure of error localization.

%\gd{fix up this notation to indicate whether there are multiple core facts that get included, ...}
% satisfying $e(\mathbf{c}_{i,k},\mathbf{c}_b) = \mathrm{not\_supported}$ where $\mathbf{c_j}$ is the \textit{banned fact}
By construction of $D'$, $\mathbf{d}_i$ is supported by at least two facts, its core atomic fact and auxiliary atomic fact(s). From this set of auxiliary atomic fact(s), we sample a \textit{banned fact} $\mathbf{c}_b$. For each $\mathbf{d}_i$, we sample a set of \textit{key facts} $C_i = \{\mathbf{c}_{i,1},\ldots,\mathbf{c}_{i,m}\}$ such that $C_i$ contains the all atomic facts of the response  $r$ except $\mathbf{c}_b$. We then prompt the LLM to generate an evidence article supporting the facts $C_i$ and not supporting the fact $c_j$. Each of these evidence articles ideally should support \textit{all} the key facts and not support the \textit{banned fact}.

% \gd{move this filter to appendix} 

%This filters out 28.3\% of the total claims. We then prompt the LLM to generate an evidence article supporting the facts $C_i$ and not supporting the fact $c_j$. Each of these evidence articles ideally should support \textit{all} the key facts and not support the \textit{banned fact}.
% By construction of $D'$, $\mathbf{d}_i$ is supported by at least two facts, it's core atomic fact and auxiliary atomic fact(s). From this set of auxiliary atomic fact(s), we sample a \textit{banned fact}, let's call it $\mathbf{c}_j)$. For each $\mathbf{d}_i$, we sample a set of facts $C_i = \{\mathbf{c}_{i,1},\ldots,\mathbf{c}_{i,m}\}$ such that $C_i$ contains the all atomic facts that are entailed by $\mathbf{d}_i$, excluding the \textit{banned fact} $\mathbf{c}_j$. It also contains additional atomic facts from the overall generation $r$. We constrain $C_i$ to \emph{not} contain a fact $\mathbf{c}_j$ such that $e(\mathbf{d}_i,\mathbf{c}_j) = \mathrm{supported}$. We will then prompt an LLM to generate an evidence article supporting the facts $C_i$. Each of these evidence articles ideally should support \textit{all} the key facts and not support the \textit{banned fact}.
%We use the map $D'$ generated in Step 3 to guide our controlled generation of synthetic evidence. We sample a set of key facts (this set always contains the \textit{core key fact} of the decontextualization) and one \textit{banned fact} from the mapping of each decontextualization set ${d}_i$.
%
The prompt for this step is detailed in Figure \ref{fig:gen_prompt} and other filtering criteria are described in Appendix \ref{appendix:filtering}. Denote this set where evidence generation is feasible as $E'$. %, which results in filtering of 4.52\% of cases.

% \ag{im not sure how to represent these numbers, (to discuss)} \gd{describe the results of this dataset construction in 1-2 sentences: explain what the figure shows}

% Finally, we are able to generate evidence for 
%We implemented the controlled evidence-generation process leading to
% 74 multi-atomic decontextualized facts. We call this set $E'$. %\jl{Minor, but maybe it'd be better to move this paragraph to earlier, having earlier high-level expectations is shown to yield shorter reading time and better comprehension rates} \ag{Noted, will leave here for now, and change the flow for arxiv version later}

\subsection{Evaluation Criteria} 

We evaluate the impacts of loss of minimality on the recall of fact-checking. We measure the percentage of cases that change their label from \texttt{\small{SUPPORTED}} to \texttt{\small{NOT\_SUPPORTED}} after decontextualization on the set $E'$. We employ the \texttt{\small{roberta-large}} from AlignScore \cite{zha2023alignscore} as our $\mathrm{Check()}$ function.\footnote{We conducted preliminary analysis with GPT-4 as well, and found it gave very similar results.} %Specifically, we evaluate each of the \textit{core key fact}, \textit{decontextualization}, and the \textit{banned fact} with the generated evidence using .
Using $\mathrm{Check(D_i, c_i)}$, we identify cases where the \textit{core key fact} is \texttt{\small{SUPPORTED}} by the generated evidence while the \textit{decontextualization} and \textit{banned fact} are \texttt{\small{NOT\_SUPPORTED}}. We call this set \emph{auto non-minimal}.
%Our goal is to determine the impacts of minimality on error localization and approximate an upper bound for the occurrence of these errors in a dataset. 

%\section{\colorbox[HTML]{DCE5D2}{Results B: Minimality}}
% .\gd{discuss this data source earlier} 

% We find that 33.95\% of the decontextualizations\gd{this doesn't align to a table? or does it? if this is setup, it should occur before the ``results discussion'' section, in 7.0} \ag{yes, sorry my bad} only match to a single atomic claim and thus do not pose the minimality loss issue. We also discard the cases where the banned fact is too similar to the key facts chosen for generation, for example, the key fact and banned fact pair (\textit{`The magnetic fields of neutron stars exist.', `Neutron stars have magnetic fields.'}) is not considered for controlled evidence generation as they are closely related. This constitutes 28.31\% cases. 

\begin{table}[t!]
\small
    \centering
    \begin{tabular}{ccc}
        \toprule

         \textbf{Baseline} & \textbf{Potential} & \textbf{Auto}\\
         \textbf{} & \textbf{ Non-minimal} & \textbf{Non-minimal}\\
        \midrule
        %\multirow{3}{*}{} & Accuracy & XX\% \\
         %16.97\% &  9.21\%\\
         % OLD & 12.39\% & 9.21\% \\
         \texttt{\small{SAFE-DECONTEXT}} & 8.49\% & 3.94\% \\
         \texttt{\small{SIMPLE-DECONTEXT}} & 23.39\% & 13.42\%\\
        % 12 / 48 cases where key_fact=true, decontext=False
        % 104 / 597
        % 
        \bottomrule
    \end{tabular}
    % \caption{Fraction of dataset that gets impacted by minimality loss and fraction of the entire dataset that leads to errors. }
    \caption{Percentage of overall dataset impacted by minimality loss due to decontextualization leading to prediction changes from \texttt{\small{SUPPORTED}} to \texttt{\small{NOT\_SUPPORTED}}.}

    \label{tab:minimality}
\end{table}

\subsection{Results}
Table~\ref{tab:minimality} shows the fraction of claims which are included in the set $E'$, which yields 8.49\% for \texttt{\small{SAFE-DECONTEXT}} and 23.39\% for \texttt{\small{SIMPLE-DECONTEXT}}. We refer to these claims as \emph{potential non-minimal claims}: they have passed the checks in our pipeline and contain multiple atomic facts. Next we apply the $\mathrm{Check()}$ function to identify auto non-minimal claims, and find that they occur at a rate of 3.94\% to 13.42\% (Table \ref{tab:minimality}). %This highlights the sensitivity of the rate of occurrence of multi-atomic decontextualization to the decontextualization method used. We further analyze these with a human study.
%However, we don't yet know whether they really represent different pieces of information that may or may not be contained in an evidence paragraph.

% \gd{I think this sentence is a little confusing. is it correct to say: these pass all the checks, but we want to confirm that they really are non-atomic?}

%We analyze results generated set $E'$ which consists generations from 16.97\% of the claims in FactBench dataset.\gd{unclear where this comes from: 100 - 33.95 - 28.3 is not it?} These are the cases which are susceptible to error localization due to loss of minimality. 
% \gd{unify with earlier notation} 

%This result shows that nearly 10\% of the decontextualized claims contain multiple facts. %This shows a high degree of potentially non-minimal claims.

%shows that for the FactBench dataset, we can generate evidence articles that impact the supportedness decision for 9.21\% of the dataset.
%This represents the cases out of total claims where minimality loss due to decontextualization leads to loss of error localization.

% \begin{table}[t!]
% \renewcommand{\tabcolsep}{1.5mm}
% \centering
% \small
% \begin{tabular}{c|c}
% \toprule
% \textbf{Category} & \textbf{Percentage Occurances} \\
% \midrule
% \texttt{Minimal}   & 30.90\% \\
% \texttt{Non-minimal} & 69.10\% \\
% \bottomrule
% \end{tabular}
% \vspace{-2mm}
% \caption{Human annotation for categorizing auto non-minimal subset into minimal vs.~non-minimal.}
% \label{tab:min_nonmin}
% \vspace{-3mm}
% \end{table}

\begin{table}[t!]
\renewcommand{\tabcolsep}{1.5mm}
\centering
\small
\begin{tabular}{c|c|c}
\toprule
\textbf{Category} & \textbf{Minimal}  & \textbf{Non-minimal}\\
\midrule
\texttt{\small{SAFE-DECONTEXT}}  & 56.2\% & 43.8\% \\
\texttt{\small{SIMPLE-DECONTEXT}} & 27.5\%  & 72.5\% \\
\bottomrule
\end{tabular}
\vspace{-2mm}
\caption{Human annotation for categorizing the Auto Non-minimal subset into minimal vs.~non-minimal.}
\label{tab:min_nonmin}
\vspace{-3mm}
\end{table}

% 
% \begin{table}[t!]
% \renewcommand{\tabcolsep}{1.5mm}
% \centering
% \small
% \begin{tabular}{c|c}
% \toprule
% \textbf{Decontextualization Edit Type}  & \textbf{Percentage} \\
% \midrule
% \texttt{Additional Information} & 69.09\% \\
% \texttt{Pronoun Swap}            &  5.45\% \\
% \texttt{Bridging}                & 10.91\% \\
% \texttt{Global Scoping}          &  7.27\% \\
% \texttt{Bad Decomposition}       &  7.27\% \\
% \bottomrule
% \end{tabular}
% \vspace{-2mm}
% \caption{Categorizing the type of Information added by Decontextualization to Auto Non-Minimal Subset.}
% \label{tab:add_error_info}
% \vspace{-3mm}
% \end{table}

\begin{table*}[t]
\renewcommand{\tabcolsep}{1.5mm}
    \centering
    \small
    \begin{tabular}{c|c|c|c|c|c}
    \toprule
       {\multirow{2}{*}{}} & {\texttt{\small{ACCURACY}}} & {\texttt{\small{ACCURACY}}} & {\texttt{\small{ACCURACY}}} &  \texttt{\small{MODIFICATION}}& {\texttt{\small{AVG LENGTH}}}\\
       \textbf{Subset} & {\texttt{\small{OVERALL}}} & {\texttt{\small{SUPPORTED}}} & {\texttt{\small{NOT\_SUPPORTED}}} & \texttt{\small{RATE}} & {\texttt{\small{\textit{(\# of words)}}}}\\ 
       \midrule
        \texttt{\small{ATOMIC}} & \cellcolor{red!20}68.7\% & \cellcolor{red!20}77.5\% &  \cellcolor{red!20}22.4\% & - & 7.61$\pm$3.03\\
        % \texttt{\small{DECONTEXT}} & 72.3 & 80.3 &  30.2 \\
        % \texttt{\small{SAFE\_DECONTEXT}} & 72.7 & 81.0 & 29.3  \\
        \texttt{\small{SIMPLE-DECONTEXT}} & \cellcolor{green!20}76.2\% & \cellcolor{green!20}84.3\% & 33.6\% & 99.5\% &15.55$\pm$5.65\\
        \texttt{\small{SAFE-DECONTEXT}} & 73.4\% & 81.3\% & 31.9\% & 72.6\% & 9.86$\pm$4.38 \\
        \texttt{\small{MOLECULAR-DECONTEXT}} & 74.7\% & 81.5\% & \cellcolor{green!20}38.8\% & 96.8\% & 14.96$\pm$5.6 \\
        % \texttt{\small{MOLECULAR}} & 72.9 & 78.5 & 43.1  \\
        % \texttt{\small{MOLEC\_GPT4\_OLD}} & 74.5 & 81.3 & 38.8  \\ 
        
    \bottomrule
    \end{tabular} \vspace{-2mm}
    \caption{Accuracy measured by \(\mathrm{Check}(D_{i}, m)\), assessing the effectiveness of claim revisions by each baseline against the ambiguous document set associated with claim's main entity. }
    \label{tab:exp1_results}
    \vspace{-3mm}
\end{table*}

\begin{table*}[t!]
\renewcommand{\tabcolsep}{1.5mm}
    \centering
    \small
    \begin{tabular}{c|c|c|c|c|c}
    \toprule
       {\multirow{3}{*}{}\textbf{Human Label$\to$}} & \multicolumn{3}{c|}{\texttt{\small{SUPPORTED}}} & \texttt{\small{NOT\_SUPPORTED}} & {\multirow{3}{*}{}}\\
       \cmidrule(lr){2-4} \cmidrule(lr){5-5}
       \textbf{Baseline Pred$\to$} & {\texttt{\small{SUPPORTED}}} & {\texttt{\small{SUPPORTED}}} & {\texttt{\small{NOT\_SUPPORTED}}} & \texttt{\small{SUPPORTED}}\\
       \cmidrule(lr){2-5} 
       \textbf{Matching Type$\to$}
        &  Multi-Evidence & Single-Evidence & No Evidence & Single/Multiple & \textbf{Overall}\\
        \textbf{   Baseline $\downarrow$} & matched & Wrong Entity & matched  & Evidence matched  & $\downarrow$ \\
        \midrule
        \texttt{\small{ATOMIC}} & \cellcolor{red!20} 16.2\% &  \cellcolor{green!20}0.8\% & \cellcolor{green!20} 1.8\% & \cellcolor{red!20} 12.4\% & \cellcolor{red!20} 31.1\%\\
        \texttt{\small{SIMPLE-DECONTEXT}} & \cellcolor{green!20} 7.9\%  & \cellcolor{red!20} 1.5\% & 3.9\% & 10.6\% & \cellcolor{green!20} 23.8\%\\
        \texttt{\small{SAFE-DECONTEXT}} & 12.0\%  & 1.0\% & \cellcolor{green!20} 2.8\% & 10.9\% & 26.6\%\\
        \texttt{\small{MOLECULAR-DECONTEXT}} & 9.2\% & \cellcolor{red!20} 1.5\% & 4.8\% & \cellcolor{green!20} 9.8\% & 25.3\%\\
    \bottomrule
    \end{tabular} \vspace{-2mm}
    \caption{Fine-grained error analysis categorizing baseline mistakes based on human label of \texttt{ \small SUPPORTED/NOT\_SUPPORTED} along with categorization of \texttt{\small <Single/Multi/No>-Evidence} based on the number of ambiguous evidence docs that support the claim.}
    \label{tab:exp1_error_analysis}
    \vspace{-3mm}
\end{table*}

% \begin{table}[t!]
% %\renewcommand{\tabcolsep}{1.2mm}
% \small
%     \centering
%     \begin{tabular}{cccc}
%         \toprule

%          \textbf{Human Annotation} & \texttt{\small{SIMPLE}} & \texttt{\small{SAFE}} & \texttt{\small{MOLECULAR}}\\
%         \midrule
%         %\multirow{3}{*}{} & Accuracy & XX\% \\
%          %16.97\% &  9.21\%\\
%          % OLD & 12.39\% & 9.21\% \\
%          \texttt{\small{Minimal}} $\uparrow$ & 16.7\% & 25.0\% & \cellcolor{green!20}54.2\% \\
%          \texttt{\small{Non-Minimal}} $\downarrow$ & \cellcolor{red!20}50.0\% & 0.0\%& 20.8\%\\
%          \texttt{\small{Ambiguous}}  $\downarrow$ & 33.3\% & \cellcolor{red!20}75.0\% & 25.0\%\\

%         \bottomrule
%     \end{tabular}
%     % \caption{Fraction of dataset that gets impacted by minimality loss and fraction of the entire dataset that leads to errors. }
%     \caption{Human Analysis of Decontextualized Claims for all baselines on the axis of minimality and ambiguity.}

%     \label{tab:human_minimality}
% \end{table}

\subsection{Human Evaluation}

% \ag{Analysing the cases into buckets to comment on in how many of these the information added is necessary v/s loss of minimality. \url{https://docs.google.com/spreadsheets/d/13c5q-IMYa28l19dvUFBscFSzUyv2pDUpUfde4f7wFPw/edit?usp=sharing }}
\paragraph{Susceptibility to Error Localization} We perform human evaluation  on the \textit{auto non-minimal} claims in Table~\ref{tab:minimality}. First, we categorize these into human judgments of whether a claim in this subset is minimal or not in Table~\ref{tab:min_nonmin}. We categorize a decontextualization as minimal based on the criteria outlined in \ref{sec:desiderata}. This annotation is performed by the authors of the paper. We find that for \texttt{\small{SAFE-DECONTEXT}}, 43.8\% of these cases are truly non-minimal in our judgment which represent 1.7\% of the dataset $D$. For the \texttt{\small{SIMPLE-DECONTEXT}} baseline, we find that a staggering 72.5\% of the auto non-minimal subset represents truly non-minimal claims. This represents 9.6\% of the dataset $D$. We note that the remaining fraction of decontextualization cases not identified by the auto methods are those which entail more than one atomic fact but it is a necessary addition to make the atomic claim standalone.

% For example, the atomic claim \textit{``The information is from Eisenberg \& Kauzmann''} is decontextualized to \textit{``The information about water molecules being highly polar is from Eisenberg \& Kauzmann''}, a necessary addition to ensure clarity.\gd{can we merge this with the ``All taxes'' example? they are similar kinds of things} 

%We find that 69.1\% of these cases are truly non-minimal in our judgment which represent 6.36\% of the complete dataset. The remaining 31\% are decontextualization cases which entail more than one atomic fact but it is a necessary addition to make the atomic claim standalone. For example, the atomic claim \textit{``The information is from Eisenberg & Kauzmann''} is decontextualized to \textit{``The information about water molecules being highly polar is from Eisenberg & Kauzmann''}, a necessary addition to ensure clarity. 
\textbf{Decontextualization and Loss of Minimality}
We highlight that addition of information to a claim does not always make it less entailed to the evidence. In fact, in many cases information addition makes the sentence more specific. This is evident from Table \ref{tab:min_nonmin} which shows that automatically flagged cases for non-minimality have a large percentage of minimal claims after human evaluation. For instance,  ``\emph{All taxes must be paid by April 15}'' $\rightarrow$ ``\emph{In the US, all taxes must be paid by April 15}'' is a necessary addition for claim specificity. 

%\greg{say something}. %decontextualization leading to prediction change of \texttt{\small{SUPPORTED}} to \texttt{\small{NOT\_SUPPORTED}} is due to the claims lacking minimality. 

% \paragraph{Decontextualization and Loss of Minimality} 
% We further perform an analysis on the \textit{auto non-minimal} claims in Table~\ref{tab:minimality} to give them fine-grained categories adopting from the taxonomy proposed in \citet{choi2021decontextualization} shown in Table \ref{tab:add_error_info}. Additionally, we also include \texttt{\small{Bad Decomposition}} as a category to indicate that the original atomic claim was too short, or simply not check worthy. Although a large fraction of edit types performed by decontextualization fall under pronoun swapping, bridging, and global scoping which are usually necessary for the claim to be complete and minimal, most of the claims have the edit type of additional information which are also non-minimal. We highlight that addition of information to a claim does not always make it less entailed to the evidence. In fact, in many cases information addition makes the sentence more specific. For instance,  ``\emph{All taxes must be paid by April 15}'' $\rightarrow$ ``\emph{In the US, all taxes must be paid by April 15}'' is a necessary addition to make the claim more specific. 

\label{sec:conclusion_minimilaity}
\subsection{Conclusion: Problem of Non-minimality}
We find through our controlled experiment and human evaluation that decontextualization can lead to non-minimal cases for between 1.7\% to 9.6\% of decontextualizations. These cases could cause error localization issues due to too much information added to the claims. In absolute terms, this is a low fraction for the baseline \texttt{\small{SAFE-DECONTEXT}}. However, we note that a biography from FActScore \cite{min-etal-2023-factscore} contains dozens of atomic facts, meaning that in a single response from an LLM, there can easily be a handful of facts posing localization problems. Given the increasing adoption of the decomposition and decontextualization pipeline for automatic fact verification systems, we argue that multiple localization errors per response is cause to re-examine that pipeline. Next,  we analyze tradeoffs between minimality and decontextuality for fact checking of ambiguous biographies.

\label{sec:setup}
\section{Experiment: Ambiguous Biographies}
\label{sec:expta}

We now analyze to what extent our molecular facts add the correct information to decontextualize on an existing dataset with ambiguous entity references. %We focus in particular on evaluating our criterion of decontextuality: whether enough information has been added to specify a claim. We compare the performance of baselines described in section \ref{sec:baselines}.

\paragraph{Dataset} We use the ambiguous biographies dataset introduced in \citet{chiang2024merging} which comprises biographies generated by LLMs for multiple entities that share similar names, such as \textit{Dick Hanley (swimmer) }and \textit{Dick Hanley (footballer)}. 
% Although these biographies are comprised of factual claims, the aggregation of data often results in narratives that are not factually coherent due to the ambiguity stemming from entities with overlapping identifiers. 
%The dataset consists of: (i) atomic claims that require decontextualization (such as entity specification, noun completion), (ii) multiple entities with the same name that require additional disambiguation such as specifying location, occupation, or time-period, (iii) human judgments of fact verification against evidence retrieved for each disambiguated entity. 
In this dataset we represent the biographies generated by the LLMs as $\mathbf{r}$ and $\mathbf{c}_i$ correspond to atomic claims generated using the methodology outlined in \cite{min-etal-2023-factscore}. For this setting, we define each claim to have a subject $\mathbf{s_i}$, which is ambiguous due to the nature of the dataset. The dataset provides a set of evidence documents sourced from Wikipedia page of the subject disambiguation, \( D_{i} = \{D_{i,2}, D_{i,2}, \ldots\} \) for subjects sharing similar names as $\mathbf{s}_i$. This dataset is suitable for evaluating \textit{decontextuality} as it consists of two properties: (i) atomic claims that require decontextualization (such as entity specification, noun completion), (ii) multiple entities with the same name that require additional disambiguation such as specifying location, occupation, or time-period. 
Our goal is to verify the claims with the set of documents using  $\mathrm{Check()}$. We randomly sample 726 claims from the human-annotated set for this study which belong to either \texttt{\small SUPPORTED} or \texttt{\small NOT\_SUPPORTED} categories. For each claim we construct a revision using the methods and baselines described in section \ref{sec:methodology} and compare the prediction with human labels.
%Following past work \cite{min-etal-2023-factscore}, we implement $\mathrm{Check()}$ by prompting \texttt{\small \small gpt3.5-turbo} using \textit{`<evidence> <atomic fact> True or False?}'. 

\paragraph{Evaluation Criteria}

%We evaluate across 726 claims sampled from the ambiguous biography dataset.  This dataset comprises claims and ambiguous evidence gathered for entities sharing similar names with the claim's subject. Our evaluation involves making factuality verification judgments for each claim against every pair of ambiguous evidence independently. For instance a claim about \textit{Ann Jansson} will be checked against entities with similar names such as (\textit{Ann Jansson - Swedish footballer, Ann Jansson - Swedish racewalking athlete, $\ldots$}). A judgment is deemed accurate when it aligns with human annotations (either support or no support), and, in the case of support, when it correctly identifies the relevant evidence. Specifically, 
%\gd{change to ci} 
We evaluate our judgment of a claim on two axes: (1) whether it aligns with the human annotation of \texttt{\small SUPPORTED} or \texttt{\small NOT\_SUPPORTED}, and (2) whether it is supported by the correct evidence. For each evidence associated with the claim, we compute  \({p}_{i,k} =\mathrm{Check}(D_{i,k}, c_i )\) where $c_i$ is the claim processed by the particular baseline and $k$ represents the ${k}$th ambiguous subject related document for the claim. We consider the judgment \({p}_{i,k}\) to be correct only if the prediction of the claim matches the human label \textit{and} the prediction is supported by the correct entity's evidence document.  %Ideally, methods that integrate context and disambiguation should should achieve higher accuracy on \mathrm{Check()} than simple atomic claims that lack context.

\begin{table}[t!]
\small
    \centering
    \begin{tabular}{cccc}
        \toprule

         \textbf{Baseline} & \texttt{\small{Minimal}} $\uparrow$ & \texttt{\small{Non-Minimal}}$\downarrow$ & \texttt{\small{Ambig.}}$\downarrow$ \\
        \midrule
        %\multirow{3}{*}{} & Accuracy & XX\% \\
         %16.97\% &  9.21\%\\
         % OLD & 12.39\% & 9.21\% \\
         \texttt{\small{SIMPLE}}  & 16.0\% & \cellcolor{red!20}56.0\%  & 28.0\% \\
         \texttt{\small{SAFE}}  &  24.0\%   & 0.0\% & \cellcolor{red!20}76.0\% \\
         \texttt{\small{MOLECULAR}} & \cellcolor{green!20}52.0\% & 24.0\% & 24.0\%\\

        \bottomrule
    \end{tabular}
    % \caption{Fraction of dataset that gets impacted by minimality loss and fraction of the entire dataset that leads to errors. }
    \caption{Human analysis of decontextualized claims for all baselines on the axis of minimality and ambiguity.}

    \label{tab:human_minimality}
\end{table}
\section{Results: Ambiguous Biographies}
Table~\ref{tab:exp1_results} presents the results of this experiment. All methods of decontextualization baselines yield higher accuracy rates compared to atomic claims, across all subsets.
We see that Molecular and Simple decontextualization methods have a higher proclivity to modify the atomic claims than the SAFE decontextualization baseline. Consequently, the average sentence lengths of the former methods is also larger than the SAFE baseline. Higher degrees of modification generally lead to higher accuracy. All three methods are on a Pareto frontier of length versus accuracy. %\gd{I honestly don't know what to say about Molecular here...it doesn't look like the method one would prefer} \ag{The length is higher for molecular because of disambiguation in this dataset (which is necessary for almost all claims)}

However, accuracy using the $\mathrm{Check()}$ function does not incorporate minimality. We investigate the minimality of the baselines by performing a human evaluation of randomly sampled 25 claims in Table~\ref{tab:human_minimality}. We see that the baseline \texttt{\small{SIMPLE-DECONTEXT}} has a large fraction of non-minimal and ambiguous claims as compared to \texttt{\small{MOLECULAR-DECONTEXT}}. Analysis in Section \ref{sec:conclusion_minimilaity} shows that \texttt{\small{SAFE-DECONTEXT}} is more minimal than \texttt{\small{SIMPLE-DECONTEXT}}; however, it struggles with ambiguity.

\textbf{Overall, we observe that molecular claims strike a balance by maintaining minimality with ambiguity removal and improving accuracy.} They are significantly more minimal than \texttt{\small{SIMPLE-DECONTEXT}} and more performant in ambiguous generations than \texttt{\small{SAFE-DECONTEXT}}.

%The results for the experiment in Section \ref{sec:expta} are presented in 
% Table~\ref{tab:exp1_results} presents the results of this experiment. The results suggest that both basic and molecular decontextualization baselines yield higher accuracy rates compared to atomic claims, across all subsets. 

\paragraph{Error breakdown} To analyze the nature of errors encountered, we detail a case-wise error distribution in Table \ref{tab:exp1_error_analysis}. Specifically, we study the behavior of various baselines to mispredict the label as \texttt{\small{SUPPORTED}} or \texttt{\small{NOT\_SUPPORTED}} in comparison to human annotation. Note that due to the ambiguous nature of this dataset, claims may be erroneously validated by several distracting pieces of evidence. Therefore, we further partition the error analysis table to reflect the model's prediction on (i) \texttt{Single/Multi/No Evidence}: whether a claim is supported by single, multiple, or no pieces of evidence, and (ii) \texttt{(Correct/Wrong Entity)}: whether the set of supporting evidence contains the accurate evidence with which the claim ought to be aligned. Overall, all decontextualization methods show a lower error rate than atomic claims. %We discuss the specific cases in the following section.  

\begin{table}[t!]
\renewcommand{\tabcolsep}{1.5mm}
\centering
\small
\begin{tabular}{c|c}
\toprule
\textbf{Baseline Pair} & \textbf{Overlap}  \\
\midrule
\texttt{\small{ATOM \& SIMPLE-DECONTEXT}} & 7\%  \\
\texttt{\small{ATOM \& SAFE-DECONTEXT}}  & 44\%\\
\texttt{\small{ATOM \& MOLECULAR-DECONTEXT}} & 15\% \\
\texttt{\small{SIMPLE-DECONTEXT \& SAFE-DECONTEXT}} & 27\%  \\
\texttt{\small{SIMPLE-DECONTEXT \& MOLECULAR-DECONTEXT}} & 36\%  \\
\texttt{\small{MOLECULAR-DECONTEXT \& SAFE-DECONTEXT}} & 32\%  \\

\bottomrule
\end{tabular}
\vspace{-2mm}
\caption{Information overlap between baselines as measured by bi-directional entailment.}
\label{tab:entailment}
\vspace{-3mm}
\end{table}

\paragraph{Information Overlap} We perform an information overlap analysis shown in Table \ref{tab:entailment} using the model from \citet{liu2022wanli} to check bidirectional entailment of the fraction of cases where the information is equivalent between two baselines \cite{gunjal2023drafting}. We find in a large fraction of cases each baseline adds different information to modify the atomic claim. \texttt{\small{SAFE-DECONTEXT}} has least amount of modification albeit suffers with ambiguity and \texttt{\small{SIMPLE-DECONTEXT}} has most amount of modification at the cost of minimality loss. 

\section{Related Work}

%\paragraph{Fact Verification of LLM Generations}
Recent research in factuality verification of LLM generations advocates decomposing LLM generations into atomic facts or subclaims and verifying each against retrieved evidence \cite{min-etal-2023-factscore, kamoi-etal-2023-wice, fabbri-etal-2022-qafacteval}. End-to-end pipelines for factuality verification have been proposed, involving steps such as claim extraction, revision, determining checkworthiness, evidence retrieval, and verification \cite{wang2024factcheckbench, chern2023factool, wei2024long, chen2023complex}. These papers often evaluate on recently-released datasets of errors in generations \citet{liu2023evaluating,malaviya2023expertqa,chen2023felm}. Our work comments on the decontextualization step frequently used in these pipelines.

Our work fits into a broader ecosystem of techniques in this area. \citet{gao2023enabling} enable LLMs to generate text with citations. For faithful LLM generations, \citet{gao-etal-2023-rarr} use evidence retrieval for revision, and \citet{he2022rethinking} utilize chain-of-thought coupled with retrieval for faithful explanations. Fine-tuned systems, such as that by \citet{zha2023alignscore}, predict alignment scores for verification, while \citet{tang2024minicheck} propose LLM-AggreFact for sentence-level factuality labels. \citet{wanner2024closer} find that evaluation metrics for fact verification are sensitive to the claim decomposition method used.

%\paragraph{Decontextualization and Ambiguity} 
Prior work on decontextualization has investigated basic notions like anaphora resolution \cite{choi2021decontextualization}, question answering frameworks \cite{newman2023question}, and extract-then-decontextualize methods for summarization \cite{potluri2023concise}. In fact verification, atomic claims are made standalone before evidence retrieval via decontextualization \cite{wang2024factcheckbench} or claim revision \cite{wei2024long}. Decontextualization is also used to resolve ambiguity \citet{zhang2021situatedqa,lee2024ambigdocs}; our work shares this focus.

\section{Conclusion}
We introduce molecular facts and the desiderata of decontextualization in LLM fact verification. We define the criteria of decontextuality and minimality in this context. %two criteria important for modifying atomic facts: (i) Decontextualization with unique specification of entities; and (ii) ensuring Minimality of edits. 
Through a controlled experiment, we show that localization errors due to loss of minimality by decontextualization is sensitive to the method used. We propose a method of ``molecular facts'' and find that they improve fact verification precision for claims from generation about ambiguous entities. We show that molecular facts strike a balance between maintaining minimality and accuracy of fact-verification.  % Finally, we note a drastic gap between decontextualization methods and human performance in cases of ambiguous claim verification. 

% In this work, we illustrate the limitations of atomic fact verification for long-form generations. We define the concept of a ``molecular fact,'' expanding on atomic facts to fully decontextualize and disambiguate entities and other references. We conduct experiments to show that incorporating information for decontextualization can lead to more reliable fact-checking in cases of ambiguous entities, while still allowing for localization due to the minimality of molecular facts. 

\section*{Limitations}
\paragraph{Scope} We illustrate the phenomenon of ambiguity in atomic claims; however, our main evaluation of molecular facts is in the domain of English-language biographies. This is due to the availability of the dataset, Wikipedia evidence, and the prevalence of biography benchmarks in recent work. Conceptually, the ambiguity in the subject or predicate of the claim can be extended to other realistic datasets, but we leave that exploration to future work. Relatedly, we focus on entity ambiguity for illustration of our method. There may be other types of ambiguities that molecular fact generation can address in other contexts and other datasets.

Furthermore, we focus our experiments on high-performing LLMs in this work. The extension of decontextualization and molecular fact generation to smaller, open-source models and the improvement in this regime is a good subject for further study.

Finally, we believe our approach should be evaluated fully end-to-end in an LLM pipeline that generates responses and then verifies their factuality. However, despite substantial research in these directions, we are not aware of an off-the-shelf experimental pipeline that is usable for this setting.

\paragraph{Decomposition Quality} We do not consider the errors introduced due to poor decomposition of atomic facts in this work. It is possible that some of these errors are resolved due to decontextualization or disambiguation implicitly, but we do not make any specific claims about this.

\paragraph{Coverage of Domains and Languages} The datasets utilized for ambiguous biographies are limited to English-language claims focused on English-centric concepts within Wikipedia. Similarly, the synthetic data generation experiment for minimality analysis is confined to English language outputs and relies on GPT-4’s parametric knowledge, which may limit the breadth of topics and domains covered. 

\section*{Acknowledgments}
We thank Jessy Li for comments and feedback on the initial draft of this work. This work was supported by NSF CAREER Award IIS-2145280 and the NSF AI Institute for Foundations of Machine Learning (IFML). This material is also based on research that is in part supported by the Air Force Research Laboratory (AFRL), DARPA, for the KAIROS program under agreement number FA8750-19-2-1003.

\bibliography{custom}

\begin{thebibliography}{43}
\expandafter\ifx\csname natexlab\endcsname\relax\def\natexlab#1{#1}\fi

\bibitem[{Achiam et~al.(2023)Achiam, Adler, Agarwal, Ahmad, Akkaya, Aleman, Almeida, Altenschmidt, Altman, Anadkat et~al.}]{achiam2023gpt}
Josh Achiam, Steven Adler, Sandhini Agarwal, Lama Ahmad, Ilge Akkaya, Florencia~Leoni Aleman, Diogo Almeida, Janko Altenschmidt, Sam Altman, Shyamal Anadkat, et~al. 2023.
\newblock Gpt-4 technical report.
\newblock \emph{arXiv preprint arXiv:2303.08774}.

\bibitem[{Chen et~al.(2024)Chen, Kim, Sriram, Durrett, and Choi}]{chen2023complex}
Jifan Chen, Grace Kim, Aniruddh Sriram, Greg Durrett, and Eunsol Choi. 2024.
\newblock {Complex claim verification with evidence retrieved in the wild}.
\newblock In \emph{Proceedings of the North American Chapter of the Association for Computational Linguistics}.

\bibitem[{Chen et~al.(2023{\natexlab{a}})Chen, Zhao, Zhang, Chern, Gao, Liu, and He}]{chen2023felm}
Shiqi Chen, Yiran Zhao, Jinghan Zhang, I-Chun Chern, Siyang Gao, Pengfei Liu, and Junxian He. 2023{\natexlab{a}}.
\newblock \href {http://arxiv.org/abs/2310.00741} {{FELM: Benchmarking Factuality Evaluation of Large Language Models}}.
\newblock In \emph{Thirty-seventh Conference on Neural Information Processing Systems Datasets and Benchmarks Track}.

\bibitem[{Chen et~al.(2023{\natexlab{b}})Chen, Buthpitiya, Fabrikant, Roth, and Schuster}]{chen-etal-2023-propsegment}
Sihao Chen, Senaka Buthpitiya, Alex Fabrikant, Dan Roth, and Tal Schuster. 2023{\natexlab{b}}.
\newblock \href {https://doi.org/10.18653/v1/2023.findings-acl.565} {{P}rop{S}egm{E}nt: A {L}arge-{S}cale {C}orpus for {P}roposition-level {S}egmentation and {E}ntailment {R}ecognition}.
\newblock In \emph{Findings of the Association for Computational Linguistics: ACL 2023}, pages 8874--8893, Toronto, Canada. Association for Computational Linguistics.

\bibitem[{Chern et~al.(2023)Chern, Chern, Chen, Yuan, Feng, Zhou, He, Neubig, Liu et~al.}]{chern2023factool}
I-Chun Chern, Steffi Chern, Shiqi Chen, Weizhe Yuan, Kehua Feng, Chunting Zhou, Junxian He, Graham Neubig, Pengfei Liu, et~al. 2023.
\newblock {FacTool: Factuality Detection in Generative AI--A Tool Augmented Framework for Multi-Task and Multi-Domain Scenarios}.
\newblock \emph{arXiv preprint arXiv:2307.13528}.

\bibitem[{Chiang and yi~Lee(2024)}]{chiang2024merging}
Cheng-Han Chiang and Hung yi~Lee. 2024.
\newblock {Merging Facts, Crafting Fallacies: Evaluating the Contradictory Nature of Aggregated Factual Claims in Long-Form Generations}.
\newblock \emph{arXiv 2402.05629}.

\bibitem[{Choi et~al.(2021)Choi, Palomaki, Lamm, Kwiatkowski, Das, and Collins}]{choi2021decontextualization}
Eunsol Choi, Jennimaria Palomaki, Matthew Lamm, Tom Kwiatkowski, Dipanjan Das, and Michael Collins. 2021.
\newblock {Decontextualization: Making sentences stand-alone}.
\newblock \emph{Transactions of the Association for Computational Linguistics}, 9:447--461.

\bibitem[{Fabbri et~al.(2022)Fabbri, Wu, Liu, and Xiong}]{fabbri-etal-2022-qafacteval}
Alexander Fabbri, Chien-Sheng Wu, Wenhao Liu, and Caiming Xiong. 2022.
\newblock \href {https://doi.org/10.18653/v1/2022.naacl-main.187} {{QAF}act{E}val: Improved {QA}-based factual consistency evaluation for summarization}.
\newblock In \emph{Proceedings of the 2022 Conference of the North American Chapter of the Association for Computational Linguistics: Human Language Technologies}, pages 2587--2601, Seattle, United States. Association for Computational Linguistics.

\bibitem[{Falke et~al.(2019)Falke, Ribeiro, Utama, Dagan, and Gurevych}]{falke2019ranking}
Tobias Falke, Leonardo F.~R. Ribeiro, Prasetya~Ajie Utama, Ido Dagan, and Iryna Gurevych. 2019.
\newblock \href {https://doi.org/10.18653/v1/P19-1213} {Ranking generated summaries by correctness: An interesting but challenging application for natural language inference}.
\newblock In \emph{Proceedings of the 57th Annual Meeting of the Association for Computational Linguistics}, pages 2214--2220, Florence, Italy. Association for Computational Linguistics.

\bibitem[{Gao et~al.(2023{\natexlab{a}})Gao, Dai, Pasupat, Chen, Chaganty, Fan, Zhao, Lao, Lee, Juan, and Guu}]{gao-etal-2023-rarr}
Luyu Gao, Zhuyun Dai, Panupong Pasupat, Anthony Chen, Arun~Tejasvi Chaganty, Yicheng Fan, Vincent Zhao, Ni~Lao, Hongrae Lee, Da-Cheng Juan, and Kelvin Guu. 2023{\natexlab{a}}.
\newblock \href {https://doi.org/10.18653/v1/2023.acl-long.910} {{{RARR}: Researching and Revising What Language Models Say, Using Language Models}}.
\newblock In \emph{Proceedings of the 61st Annual Meeting of the Association for Computational Linguistics (Volume 1: Long Papers)}, pages 16477--16508, Toronto, Canada. Association for Computational Linguistics.

\bibitem[{Gao et~al.(2023{\natexlab{b}})Gao, Yen, Yu, and Chen}]{gao2023enabling}
Tianyu Gao, Howard Yen, Jiatong Yu, and Danqi Chen. 2023{\natexlab{b}}.
\newblock {Enabling Large Language Models to Generate Text with Citations}.
\newblock In \emph{Empirical Methods in Natural Language Processing (EMNLP)}.

\bibitem[{Goyal and Durrett(2021)}]{goyal2020annotating}
Tanya Goyal and Greg Durrett. 2021.
\newblock Annotating and modeling fine-grained factuality in summarization.
\newblock In \emph{Proceedings of NAACL}.

\bibitem[{Gunjal and Durrett(2023)}]{gunjal2023drafting}
Anisha Gunjal and Greg Durrett. 2023.
\newblock {Drafting Event Schemas using Language Models}.
\newblock \emph{arXiv preprint arXiv:2305.14847}.

\bibitem[{He et~al.(2022)He, Zhang, and Roth}]{he2022rethinking}
Hangfeng He, Hongming Zhang, and Dan Roth. 2022.
\newblock Rethinking with retrieval: Faithful large language model inference.
\newblock \emph{arXiv preprint arXiv:2301.00303}.

\bibitem[{Ji et~al.(2022)Ji, Lee, Frieske, Yu, Su, Xu, Ishii, Bang, Madotto, and Fung}]{ji2022survey}
Ziwei Ji, Nayeon Lee, Rita Frieske, Tiezheng Yu, Dan Su, Yan Xu, Etsuko Ishii, Yejin Bang, Andrea Madotto, and Pascale Fung. 2022.
\newblock Survey of hallucination in natural language generation.
\newblock \emph{ACM Computing Surveys}.

\bibitem[{Kamoi et~al.(2023{\natexlab{a}})Kamoi, Goyal, and Durrett}]{kamoi2023shortcomings}
Ryo Kamoi, Tanya Goyal, and Greg Durrett. 2023{\natexlab{a}}.
\newblock Shortcomings of question answering based factuality frameworks for error localization.
\newblock In \emph{Proceedings of the 17th Conference of the European Chapter of the Association for Computational Linguistics}, pages 132--146.

\bibitem[{Kamoi et~al.(2023{\natexlab{b}})Kamoi, Goyal, Rodriguez, and Durrett}]{kamoi-etal-2023-wice}
Ryo Kamoi, Tanya Goyal, Juan Rodriguez, and Greg Durrett. 2023{\natexlab{b}}.
\newblock \href {https://aclanthology.org/2023.emnlp-main.470} {{{W}i{CE}: Real-{W}orld {E}ntailment for {C}laims in {W}ikipedia}}.
\newblock In \emph{Proceedings of the 2023 Conference on Empirical Methods in Natural Language Processing}, pages 7561--7583, Singapore. Association for Computational Linguistics.

\bibitem[{Lee et~al.(2024)Lee, Ye, and Choi}]{lee2024ambigdocs}
Yoonsang Lee, Xi~Ye, and Eunsol Choi. 2024.
\newblock Ambigdocs: Reasoning across documents on different entities under the same name.
\newblock \emph{arXiv 2404.12447}.

\bibitem[{Li et~al.(2016)Li, O’Daniel, Wu, Zhao, and Nenkova}]{li2016improving}
Junyi~Jessy Li, Bridget O’Daniel, Yi~Wu, Wenli Zhao, and Ani Nenkova. 2016.
\newblock Improving the annotation of sentence specificity.
\newblock In \emph{Proceedings of the Tenth International Conference on Language Resources and Evaluation (LREC'16)}, pages 3921--3927.

\bibitem[{Liu et~al.(2022)Liu, Swayamdipta, Smith, and Choi}]{liu2022wanli}
Alisa Liu, Swabha Swayamdipta, Noah~A. Smith, and Yejin Choi. 2022.
\newblock \href {https://doi.org/10.18653/v1/2022.findings-emnlp.508} {{{WANLI}: Worker and {AI} Collaboration for Natural Language Inference Dataset Creation}}.
\newblock In \emph{Findings of the Association for Computational Linguistics: EMNLP 2022}, pages 6826--6847, Abu Dhabi, United Arab Emirates. Association for Computational Linguistics.

\bibitem[{Liu et~al.(2023{\natexlab{a}})Liu, Zhang, and Liang}]{liu2023evaluating}
Nelson Liu, Tianyi Zhang, and Percy Liang. 2023{\natexlab{a}}.
\newblock \href {https://doi.org/10.18653/v1/2023.findings-emnlp.467} {{Evaluating Verifiability in Generative Search Engines}}.
\newblock In \emph{Findings of the Association for Computational Linguistics: EMNLP 2023}, pages 7001--7025, Singapore. Association for Computational Linguistics.

\bibitem[{Liu et~al.(2023{\natexlab{b}})Liu, Fabbri, Liu, Zhao, Nan, Han, Han, Joty, Wu, Xiong, and Radev}]{liu-etal-2023-revisiting}
Yixin Liu, Alex Fabbri, Pengfei Liu, Yilun Zhao, Linyong Nan, Ruilin Han, Simeng Han, Shafiq Joty, Chien-Sheng Wu, Caiming Xiong, and Dragomir Radev. 2023{\natexlab{b}}.
\newblock \href {https://doi.org/10.18653/v1/2023.acl-long.228} {{{Revisiting the Gold Standard: Grounding Summarization Evaluation with Robust Human Evaluation}}}.
\newblock In \emph{Proceedings of the 61st Annual Meeting of the Association for Computational Linguistics (Volume 1: Long Papers)}, pages 4140--4170, Toronto, Canada. Association for Computational Linguistics.

\bibitem[{Louis and Nenkova(2012)}]{louis2012corpus}
Annie Louis and Ani Nenkova. 2012.
\newblock A corpus of general and specific sentences from news.
\newblock In \emph{LREC}, volume 1818, page~10. Citeseer.

\bibitem[{Malaviya et~al.(2024)Malaviya, Lee, Chen, Sieber, Yatskar, and Roth}]{malaviya2023expertqa}
Chaitanya Malaviya, Subin Lee, Sihao Chen, Elizabeth Sieber, Mark Yatskar, and Dan Roth. 2024.
\newblock {ExpertQA: Expert-curated questions and attributed answers}.
\newblock In \emph{Proceedings of the North American Chapter of the Association for Computational Linguistics}.

\bibitem[{Min et~al.(2023)Min, Krishna, Lyu, Lewis, Yih, Koh, Iyyer, Zettlemoyer, and Hajishirzi}]{min-etal-2023-factscore}
Sewon Min, Kalpesh Krishna, Xinxi Lyu, Mike Lewis, Wen-tau Yih, Pang Koh, Mohit Iyyer, Luke Zettlemoyer, and Hannaneh Hajishirzi. 2023.
\newblock \href {https://doi.org/10.18653/v1/2023.emnlp-main.741} {{FA}ct{S}core: Fine-grained atomic evaluation of factual precision in long form text generation}.
\newblock In \emph{Proceedings of the 2023 Conference on Empirical Methods in Natural Language Processing}, pages 12076--12100, Singapore. Association for Computational Linguistics.

\bibitem[{Nenkova and Passonneau(2004)}]{nenkova2004evaluating}
Ani Nenkova and Rebecca~J Passonneau. 2004.
\newblock {Evaluating content selection in summarization: The pyramid method}.
\newblock In \emph{Proceedings of the human language technology conference of the north american chapter of the association for computational linguistics: Hlt-naacl 2004}, pages 145--152.

\bibitem[{Newman et~al.(2023)Newman, Soldaini, Fok, Cohan, and Lo}]{newman2023question}
Benjamin Newman, Luca Soldaini, Raymond Fok, Arman Cohan, and Kyle Lo. 2023.
\newblock A question answering framework for decontextualizing user-facing snippets from scientific documents.
\newblock In \emph{Proceedings of the 2023 Conference on Empirical Methods in Natural Language Processing}, pages 3194--3212.

\bibitem[{Pezzelle(2023)}]{pezzelle2023dealing}
Sandro Pezzelle. 2023.
\newblock Dealing with semantic underspecification in multimodal nlp.
\newblock \emph{arXiv preprint arXiv:2306.05240}.

\bibitem[{Potluri et~al.(2023)Potluri, Xu, and Choi}]{potluri2023concise}
Abhilash Potluri, Fangyuan Xu, and Eunsol Choi. 2023.
\newblock Concise answers to complex questions: Summarization of long-form answers.
\newblock \emph{arXiv preprint arXiv:2305.19271}.

\bibitem[{Rashkin et~al.(2021)Rashkin, Nikolaev, Lamm, Collins, Das, Petrov, Tomar, Turc, and Reitter}]{Rashkin2021MeasuringAI}
Hannah Rashkin, Vitaly Nikolaev, Matthew Lamm, Michael Collins, Dipanjan Das, Slav Petrov, Gaurav~Singh Tomar, Iulia Turc, and D.~Reitter. 2021.
\newblock \href {https://api.semanticscholar.org/CorpusID:245502761} {{Measuring Attribution in Natural Language Generation Models}}.
\newblock \emph{Computational Linguistics}, 49:777--840.

\bibitem[{Roit et~al.(2023)Roit, Ferret, Shani, Aharoni, Cideron, Dadashi, Geist, Girgin, Hussenot, Keller, Momchev, Ramos~Garea, Stanczyk, Vieillard, Bachem, Elidan, Hassidim, Pietquin, and Szpektor}]{roit2023factually}
Paul Roit, Johan Ferret, Lior Shani, Roee Aharoni, Geoffrey Cideron, Robert Dadashi, Matthieu Geist, Sertan Girgin, Leonard Hussenot, Orgad Keller, Nikola Momchev, Sabela Ramos~Garea, Piotr Stanczyk, Nino Vieillard, Olivier Bachem, Gal Elidan, Avinatan Hassidim, Olivier Pietquin, and Idan Szpektor. 2023.
\newblock \href {https://doi.org/10.18653/v1/2023.acl-long.344} {Factually consistent summarization via reinforcement learning with textual entailment feedback}.
\newblock In \emph{Proceedings of the 61st Annual Meeting of the Association for Computational Linguistics (Volume 1: Long Papers)}, pages 6252--6272, Toronto, Canada. Association for Computational Linguistics.

\bibitem[{Schilder(1998)}]{schilder1998underspecified}
Frank Schilder. 1998.
\newblock An underspecified segmented discourse representation theory (usdrt).
\newblock In \emph{36th Annual Meeting of the Association for Computational Linguistics and 17th International Conference on Computational Linguistics, Volume 2}, pages 1188--1192.

\bibitem[{Tang et~al.(2024)Tang, Laban, and Durrett}]{tang2024minicheck}
Liyan Tang, Philippe Laban, and Greg Durrett. 2024.
\newblock {MiniCheck: Efficient Fact-Checking of LLMs on Grounding Documents}.
\newblock \emph{arXiv preprint arXiv:2404.10774}.

\bibitem[{Wang et~al.(2024)Wang, Reddy, Mujahid, Arora, Rubashevskii, Geng, Afzal, Pan, Borenstein, Pillai, Augenstein, Gurevych, and Nakov}]{wang2024factcheckbench}
Yuxia Wang, Revanth~Gangi Reddy, Zain~Muhammad Mujahid, Arnav Arora, Aleksandr Rubashevskii, Jiahui Geng, Osama~Mohammed Afzal, Liangming Pan, Nadav Borenstein, Aditya Pillai, Isabelle Augenstein, Iryna Gurevych, and Preslav Nakov. 2024.
\newblock \href {http://arxiv.org/abs/2311.09000} {{Factcheck-{B}ench: Fine-{G}rained {E}valuation {B}enchmark for {A}utomatic {F}act-checkers}}.

\bibitem[{Wanner et~al.(2024)Wanner, Ebner, Jiang, Dredze, and Van~Durme}]{wanner2024closer}
Miriam Wanner, Seth Ebner, Zhengping Jiang, Mark Dredze, and Benjamin Van~Durme. 2024.
\newblock {A Closer Look at Claim Decomposition}.
\newblock \emph{arXiv preprint arXiv:2403.11903}.

\bibitem[{Wei et~al.(2024)Wei, Yang, Song, Lu, Hu, Tran, Peng, Liu, Huang, Du et~al.}]{wei2024long}
Jerry Wei, Chengrun Yang, Xinying Song, Yifeng Lu, Nathan Hu, Dustin Tran, Daiyi Peng, Ruibo Liu, Da~Huang, Cosmo Du, et~al. 2024.
\newblock Long-form factuality in large language models.
\newblock \emph{arXiv preprint arXiv:2403.18802}.

\bibitem[{Wu et~al.(2024)Wu, Hu, Shi, Dziri, Suhr, Ammanabrolu, Smith, Ostendorf, and Hajishirzi}]{wu2024fine}
Zeqiu Wu, Yushi Hu, Weijia Shi, Nouha Dziri, Alane Suhr, Prithviraj Ammanabrolu, Noah~A Smith, Mari Ostendorf, and Hannaneh Hajishirzi. 2024.
\newblock Fine-grained human feedback gives better rewards for language model training.
\newblock \emph{Advances in Neural Information Processing Systems}, 36.

\bibitem[{Xu et~al.(2023)Xu, Deutsch, Finkelstein, Juraska, Zhang, Liu, Wang, Li, and Freitag}]{xu2023pinpoint}
Wenda Xu, Daniel Deutsch, Mara Finkelstein, Juraj Juraska, Biao Zhang, Zhongtao Liu, William~Yang Wang, Lei Li, and Markus Freitag. 2023.
\newblock Pinpoint, not criticize: Refining large language models via fine-grained actionable feedback.
\newblock \emph{arXiv preprint arXiv:2311.09336}.

\bibitem[{Zha et~al.(2023)Zha, Yang, Li, and Hu}]{zha2023alignscore}
Yuheng Zha, Yichi Yang, Ruichen Li, and Zhiting Hu. 2023.
\newblock \href {https://doi.org/10.18653/v1/2023.acl-long.634} {{{A}lign{S}core: Evaluating Factual Consistency with A Unified Alignment Function}}.
\newblock In \emph{Proceedings of the 61st Annual Meeting of the Association for Computational Linguistics (Volume 1: Long Papers)}, pages 11328--11348, Toronto, Canada. Association for Computational Linguistics.

\bibitem[{Zhang and Choi(2021)}]{zhang2021situatedqa}
Michael~J.Q. Zhang and Eunsol Choi. 2021.
\newblock {S}ituated{QA}: Incorporating extra-linguistic contexts into {QA}.
\newblock \emph{Proceedings of the Conference on Empirical Methods in Natural Language Processing (EMNLP)}.

\bibitem[{Zhang et~al.(2024)Zhang, Press, Merrill, Liu, and Smith}]{zhang2024how}
Muru Zhang, Ofir Press, William Merrill, Alisa Liu, and Noah~A. Smith. 2024.
\newblock \href {https://openreview.net/forum?id=FPlaQyAGHu} {{How Language Model Hallucinations Can Snowball}}.
\newblock In \emph{Forty-first International Conference on Machine Learning}.

\bibitem[{Zhang and Bansal(2021)}]{zhang-bansal-2021-finding}
Shiyue Zhang and Mohit Bansal. 2021.
\newblock \href {https://doi.org/10.18653/v1/2021.emnlp-main.531} {Finding a {B}alanced {D}egree of {A}utomation for {S}ummary {E}valuation}.
\newblock In \emph{Proceedings of the 2021 Conference on Empirical Methods in Natural Language Processing}, pages 6617--6632, Online and Punta Cana, Dominican Republic. Association for Computational Linguistics.

\bibitem[{Zhang et~al.(2023)Zhang, Li, Cui, Cai, Liu, Fu, Huang, Zhao, Zhang, Chen et~al.}]{zhang2023siren}
Yue Zhang, Yafu Li, Leyang Cui, Deng Cai, Lemao Liu, Tingchen Fu, Xinting Huang, Enbo Zhao, Yu~Zhang, Yulong Chen, et~al. 2023.
\newblock {Siren's song in the {AI} ocean: a survey on hallucination in large language models}.
\newblock \emph{arXiv preprint arXiv:2309.01219}.

\end{thebibliography}

\appendix

\section{Prompts}
\label{appendix:prompts}

\begin{figure*}
    \centering
    \includegraphics[scale=0.85,trim=90mm 0mm 100mm 20mm]{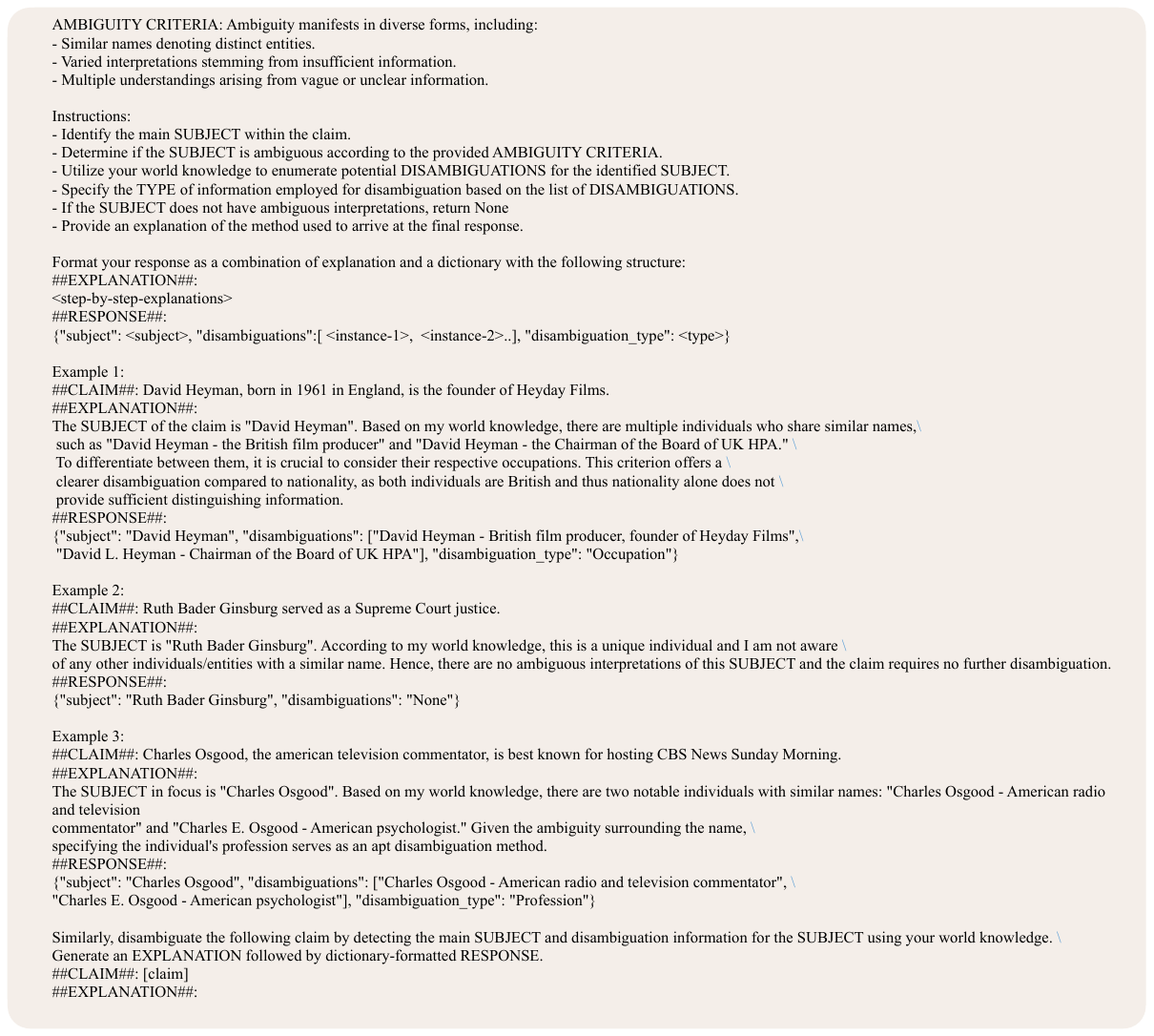}
    \caption{Ambiguity detection prompt for detection of ambiguous entities and generating disambiguation guideline for generation of molecular claims for the baselines \texttt{\small{MOLECULAR}} and \texttt{\small{MOLECULAR-GPT4}}.}
    \label{fig:ambig_prompt}
\end{figure*}

\begin{figure*}
    \centering
    \includegraphics[scale=0.90,trim=90mm 20mm 100mm 20mm]{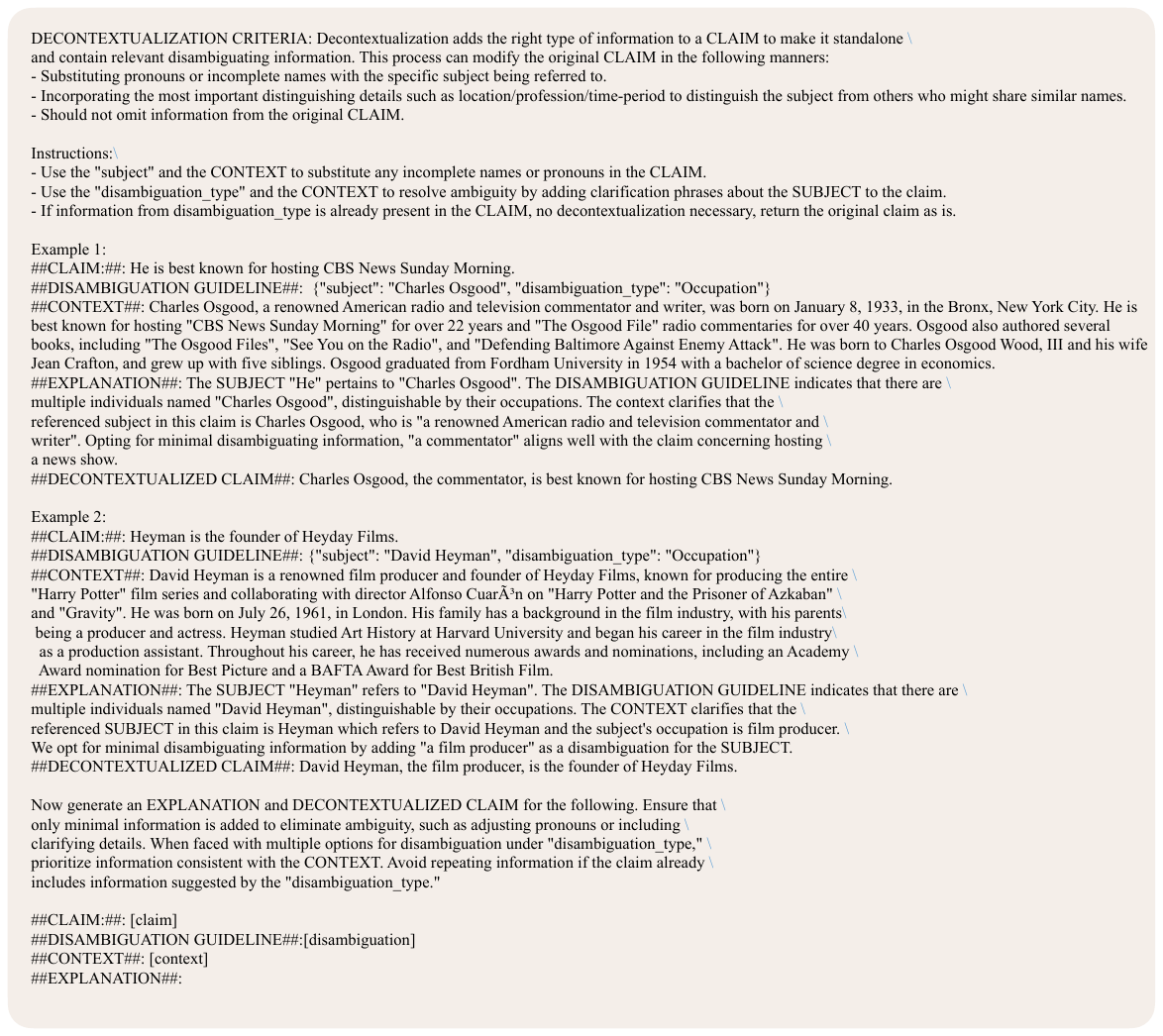}
    \caption{Molecular decontextualization prompt for the baselines \texttt{\small{MOLECULAR}} and \texttt{\small{MOLECULAR-GPT4}}.}
    \label{fig:molec_prompt}
\end{figure*}

\begin{figure*}
    \centering
    \includegraphics[scale=0.85,trim=90mm 0mm 100mm 20mm]{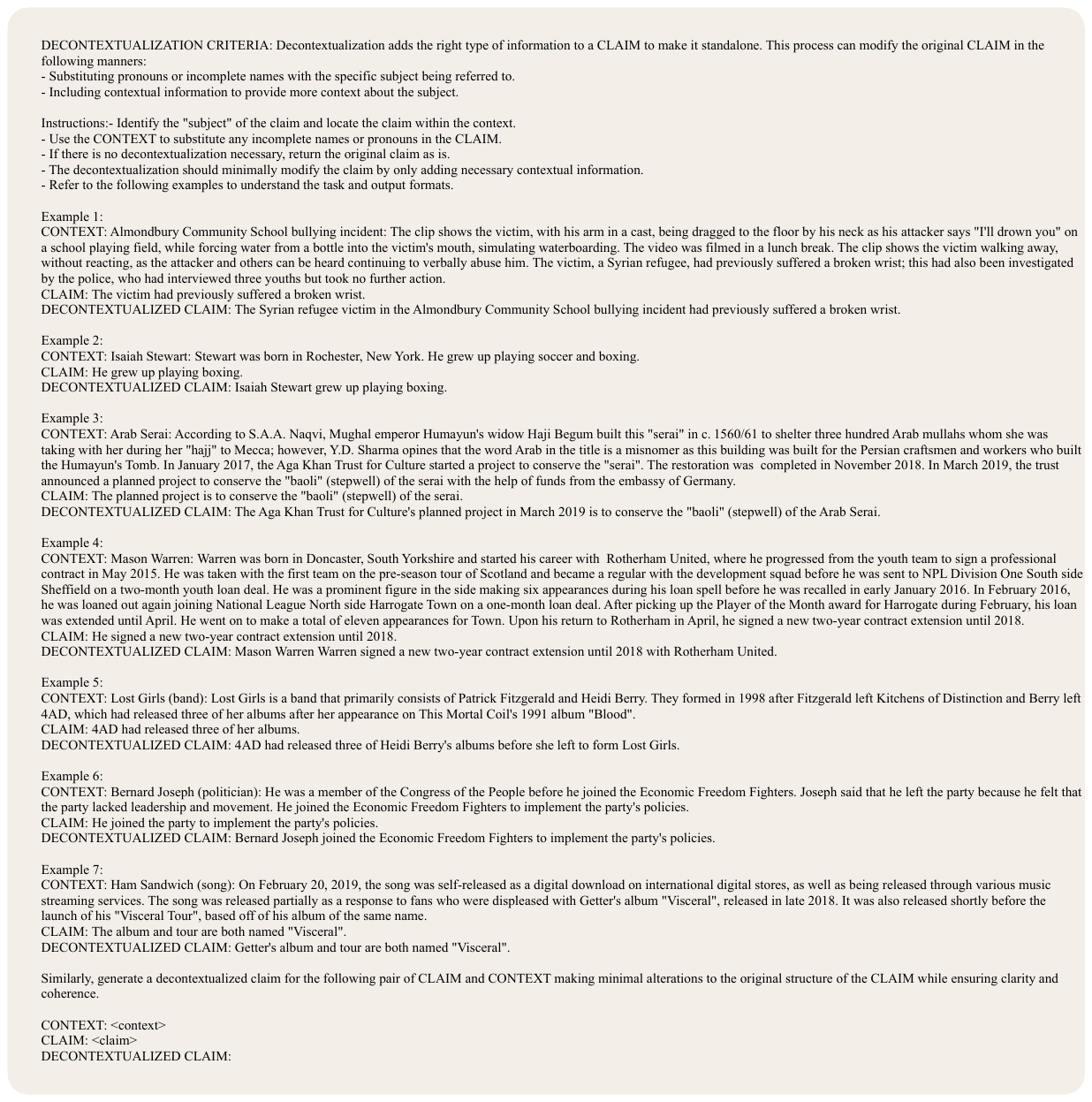}
    \caption{Decontextualization prompt for the baseline \texttt{\small{SIMPLE-DECONTEXT}}.}
    \label{fig:decontext_prompt}
\end{figure*}

We give details on all the prompts used throughout this work. 
\paragraph{Decontextuality Experiment Prompts} The step-wise molecular facts generation prompts for \texttt{\small{MOLECULAR\_DECONTEXT}} are in Figure \ref{fig:ambig_prompt}, \ref{fig:molec_prompt}. For the simple decontextualization baseline \texttt{\small{SIMPLE\_DECONTEXT}}, the prompts are provided in \ref{fig:decontext_prompt}.

\begin{figure*}
    \centering
    \includegraphics[scale=0.85,trim=90mm 170mm 100mm 20mm]{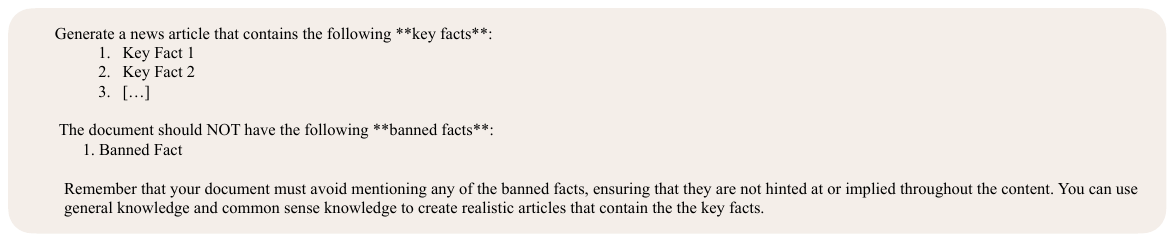}
    \caption{Prompt for controlled evidence generation to generate articles that incorporate key facts and avoid banned facts.}
    \label{fig:gen_prompt}
\end{figure*}

\paragraph{Minimality Experiment Prompts} The prompt for generating controlled evidence for the minimality experiment is given in Figure \ref{fig:gen_prompt}.

% \paragraph{Limitations of the dataset} 
% \ag{might remove this/move to appendix}
% Table \ref{tab:exp1_error_analysis} shows a relatively large error rate for cases where human annotators mark a claim as supported, whereas a subset of decontextualization methods that modify the atomic claims (\texttt{\small{decontext, molecular, gpt-molecular}}) mark it unsupported. To investigate this, we take the 178 LLM generated responses from AmbigBio dataset and prompt \texttt{\small gpt3.5-turbo} to detect the number of main entities in the passage using a simple prompt \textit{Passage: <passage>\textbackslash n\textbackslash n How many people's biography is provided in the above passage? Format your response as 'Count: <count>\textbackslash n1. <name-1> \textbackslash n2. <name2>....\textbackslash nResponse:'}. We call these silver annotations. We compare this silver entity count to the gold entity count provided by human annotators and find a disagreement in 32 / 178 = 17.97\% cases. We do a human analysis on these cases and find that 11/178 = 6.17\% of cases where the silver labels are correct. After further investigation, we find that these cases either have human annotation error or the facts about an entity are dispersed into two evidences e.g. Fact about a father is present in son's Wikipedia page and vice verse. Some other cases are when the LLMs hallucinates information about the another entity B with the same name and introduces it into the generation about entity A. \anisha{ Add Examples in appendix}.  

\section{Additional Related Work}

\paragraph{Decomposition in Text Summarization} Decomposition of responses is also prevelant in the text summarization literature. \citet{nenkova2004evaluating} introduced the Pyramid protocol for summarization evaluation which extracts weighted Summarization Content Units (SCUs) which represent the importance of various facts present in multiple human-generated summaries of a text. \citet{zhang-bansal-2021-finding} propose using Semantic Triplet Units (STUs), which are summary content units generated automatically using SRL parsers, to evaluate generated summaries with textual entailment models. Similarly, \citet{liu-etal-2023-revisiting} propose Atomic Content Units (ACUs) as a new summarization salience protocol that allows for higher inter-annotator agreement. \citet{chen-etal-2023-propsegment} propose using entailment judgments on a set of sentence propositions within a document. 

\paragraph{Decontextualization and Specificity} Decontextualization is a process of making sentences stand-alone by resolving missing context while preserving its meaning \cite{choi2021decontextualization}. A related phenomenon is the notion of \textit{specificity}. \citet{louis2012corpus} presented the first corpus of sentences distinguished on the criteria of being \textit{general} or \textit{specific}. Their idea of classification was based on examples and intuition by defining \textit{general} sentences to be broad statements about a topic that would need additional evidence or examples for a reader to understand, whereas, \textit{specific} sentences can stand by themselves. \citet{li2016improving} make this definition more specific by grounding specificity for a sentence to three requirements: (i) it is easy to understand the meaning and identify of the intended references without ambiguity; (ii) the truth of the statement can be assessed based on the sentence itself and general shared knowledge; and (iii) the sentence fully expresses key information about the participants and causes of an event. 
Another related notion is underspecification in discourse, which is an intentional feature to maintain communication efficiency \cite{schilder1998underspecified}. This has been annotated by \citet{li2016improving} and highlighted in a multimodal setting by \citet{pezzelle2023dealing}.
%, but recent work using a decontextualization step \cite{min-etal-2023-factscore, wang2024factcheckbench} has largely not engaged with these concepts.

\section{Human Annotation Criteria for Categorizing the Non-minimal Subset} We describe the criteria for annotating the auto non-minimal subset into minimal vs.~non-minimal as shown in Table \ref{tab:min_nonmin}. For each instance, we compare the original claim, the decontextualization, and the banned fact. We label cases as \textit{minimal} when either of the following applies: (1) the banned fact is closely related the atomic fact and it is a necessary addition to the atomic claim to make it standalone. In other words, the banned fact is a necessary addition to the atomic claim to add context and/or resolve ambiguity. For example, ``\emph{The album is their first full-length studio album.}'' is decontextualized to ``\emph{The album released in 2020 is Blackpink's first full-length studio album.}'' and the banned fact is ``\emph{The album was released in 2020.}''. The information in the banned fact is necessary addition to disambiguate ``\emph{the album}'' in this case. (2) The banned fact entailed by the decontextualization, but it is due to an entailment error. For example, the decontextualization \textit{``Mey Eden, one of the largest bottled water companies in Israel, offers flavored water products." }is erroneously entailed by the banned fact \textit{``Mey Eden offers still water products."}.

\section{Human Analysis Criteria for Categorizing Minimality and Ambiguity} We describe the criteria for the human analysis for on the decontextualization of each baseline on the axis of minimality and ambiguity shown in 
Table \ref{tab:human_minimality}. We categorize a claim decontextualization as \textit{non-minimal} when it contains additional information that goes beyond making the sentence stand-alone and can potentially cause loss of error-localization. We categorize a claim decontextualization as \textit{ambiguous} when it lacks clarifications for entities that could refer to different ambiguous subjects or add enough context to disambiguate the main entity. If both of the above conditions are not violated, we categorize the decontextualization as \textit{minimal}. 

\section{Models, Datasets and Computation Cost} The \texttt{\small{gpt-4-turbo-2024-04-09}} model was employed for running baselines and generating outputs, while the \texttt{\small{gpt-3.5-turbo}} model was used for evaluation through FActScore \cite{achiam2023gpt}. For generation experiments, we set the temperature to 0.75. The total cost for generating decontextualizations and evaluating the ambiguous biography experiment was approximately \$120. 

In the minimality experiment, \texttt{\small{gpt-3.5-turbo}} was used to extract atomic facts, and \texttt{\small{gpt-4-turbo-2024-04-09}} was used for decontextualization and generation tasks. This resulted in a total cost of around \$100. We use a NVIDIA A40 GPU for evaluation using AlignScore \cite{zha2023alignscore} and entailment computation using WANLI \cite{liu2022wanli},

We use ChatGPT for improving writing formatting and generating boilerplate code for figure generation in this paper.

We use the open-source dataset published by \citet{wang2024factcheckbench} under the Apache 2.0 license. We also use the open-source code-base of FactScore \cite{min-etal-2023-factscore} for evaluations which is published under MIT license and AlignScore \cite{zha2023alignscore} published under MIT License.

\begin{figure}[t]
    \centering
    \includegraphics[scale=0.32]{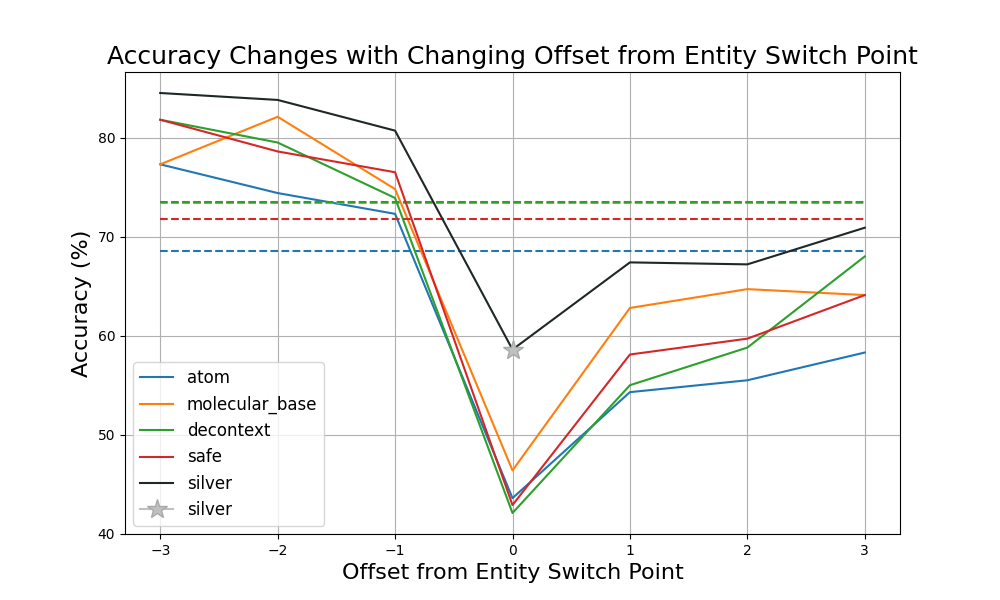}
    \caption{Variation in accuracy for different fact-checking methods as the offset from the entity switch point changes. Each line represents a method, with the solid lines indicating the method's accuracy at different offsets, and the dashed lines representing the overall accuracy of the method. The \textbf{\small{silver}} star represents the performance of human-in-the-loop molecular claim generation.}
    \label{fig:entity_switch}
\end{figure}
% \begin{table*}[t!]
% \renewcommand{\tabcolsep}{1.5mm}
% \centering
% \small
% \begin{tabular}{c|c|c|c|c|c}
% \toprule
% \textbf{Baseline} & \colorbox[HTML]{D2B48C}{\texttt{DATA\_ERR}} & \colorbox[HTML]{FFCCCC}{\texttt{DECONTEXT\_ERR}} & \colorbox[HTML]{CCCCFF}{\texttt{DISAMBIG\_ERR}} & \colorbox[HTML]{C8A2C8}{\texttt{ENTAILMENT\_ERR}} & \colorbox[HTML]{90EE90}{\texttt{NO\_ERR}} \\
% \midrule
% \texttt{DECONTEXT}       & 26.67\%  & 36.67\%   & 3.33\%   & 10.00\%   & 23.33\%   \\
% \texttt{SAFE}            & 30.00\%  & 36.67\%   & 6.67\%   & 10.00\%   & 16.67\%   \\
% \texttt{MOLECULAR}       & 33.33\%  & 3.33\%    & 23.33\%  & 23.33\%   & 16.67\%   \\
% \texttt{MOLECULAR\_GPT4} & 33.33\%  & 10.00\%   & 20.00\%  & 10.00\%   & 26.67\%   \\
% \midrule
% \texttt{OVERALL}         & 30.83\%  & 21.67\%   & 13.33\%  & 13.33\%   & 20.83\%   \\
% \bottomrule
% \end{tabular}
% \vspace{-2mm}
% \caption{Human evaluation of error types encountered in ambiguous biography generations.}
% \label{tab:human-eval}
% \vspace{-3mm}
% \end{table*}

\section{Controlled Experiment on Minimality Generation Details}
\label{appendix:filtering}
\paragraph{Filtering Criteria applied in Step 3} Before filtering claims which are supported by more than two atomic facts, we do not consider cases where one atomic fact is a substring of another one.

\paragraph{Filtering Criteria applied in Step 4} We detail the filtering criteria applied in evidence generation for partial support detailed in \ref{subsec:step4}. After we sample a set of \textit{key facts} $C_i = \{\mathbf{c}_{i,1},\ldots,\mathbf{c}_{i,m}\}$ such that $C_i$ contains the all atomic facts of the response  $r$ except $\mathbf{c}_b$, we also apply a filtering criteria to remove cases where the \textit{banned fact} and \textbf{any} of the \textit{key facts} is similar; i.e., for $\mathbf{c}_{i,k}$ in $C_i$, we filter cases where $e(\mathbf{c}_{i,k},\mathbf{c}_b) = \mathrm{supported}$. At the end of step 4 after we prompt the LLM to generate an evidence article, we also account for generation errors and remove the cases where banned fact is supported by the generated evidence.

\section{Remaining Challenges}
To shed light on the remaining challenges, we focus on one of the most challenging scenarios for decontextualization. In the ambiguous biography dataset from \citet{chiang2024merging}, we often observe what we call an \emph{entity switch point}: a claim $\mathbf{c}_i$ that draws on information about entity B, when sentences $\mathbf{c}_{<i}$ all refer to entity A. This is where decontextualization is crucial to recognize that $\mathbf{c}_i$ in context does not refer to the correct entity.

%\gd{make this a separate subsection, and it doesn't flow well}
\paragraph{Molecular claims recover fastest at the entity-switching point} We investigate the performance of baselines under the lens of ambiguity resolution. Note that these results are reported on baselines tested with \texttt{\small{gpt3.5-turbo}}. We find that the dataset of ambiguous biographies becomes the most confusing at the entity switch point. Figure \ref{fig:entity_switch} shows a significant performance drop at the switch across all methods. Basic decontextualization methods (\texttt{\small{DECONTEXT, SAFE-DECONTEXT}}) perform the worst, underperforming the \texttt{\small{ ATOMIC}} baseline at the switch, but molecular claims, which incorporate richer disambiguation information, show relative robustness, improving by 3.5\% over the most effective decontextualization approach (\texttt{\small{SAFE-DECONTEXT}}).
% \gd{commented out the rest of this paragraph, didn't seem necessary} %Unsurprisingly, the basic atomic facts method, lacking in context and disambiguation, performs the poorest. But even decontextualization and molecular methods currently struggle with correctly interpreting the facts in context and adding the right disambiguation to curtail the ambiguity. Molecular claims also show the fastest performance recovery from the entity switch point as compared to other methods.

\begin{figure}[t]
    \centering

    \includegraphics[scale=0.60,trim=35mm 140mm 90mm 0mm]{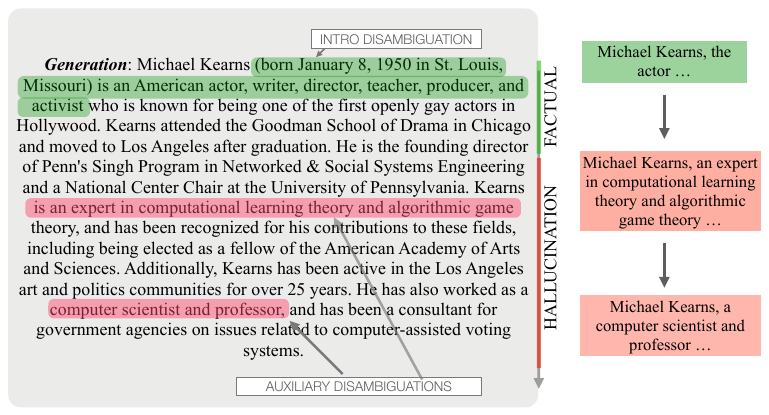}
    \caption{Changing preferences of selection of diambiguating fact by molecular decontextualization for long-form generation with hallucinations.}
    \label{fig:hallucination}
\end{figure}

\begin{figure*}
    \centering
    \includegraphics[scale=0.85,trim=90mm 80mm 100mm 20mm]{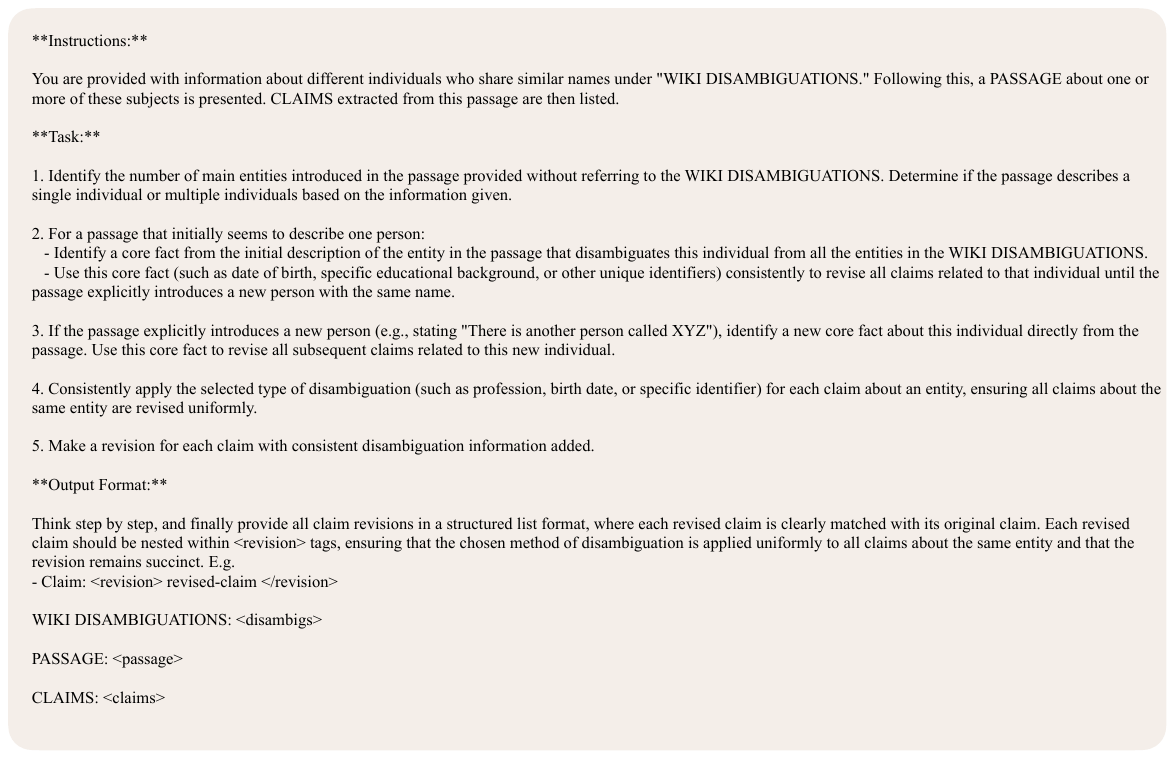}
    \caption{Silver labels ambiguity detection prompt for detection of ambiguous entities and generating disambiguation guideline for generation of molecular claims for the baselines \texttt{\small{MOLECULAR}} and \texttt{\small{MOLECULAR-GPT4}}.}
    \label{fig:silver_prompt}
\end{figure*}

\paragraph{Gap from human performance}
To estimate the upper bound of ideal performance at the entity switch point in Figure \ref{fig:entity_switch}, we generate molecular claims at the entity-switch point with weak supervision human-in-the-loop supervision. We use the prompt shown Figure \ref{fig:silver_prompt} in which has access to gold disambiguations from Wikipedia about the entities in the passage. This method's performance even with weak human supervision is significantly better than automated decontextualization methods, bringing attention to this limitation of current fact-checking pipelines.

\newpage

\end{document}